\documentclass[journal,twoside]{IEEEtran}
\IEEEoverridecommandlockouts
\usepackage{cite}
\usepackage{amsmath,amssymb,amsfonts}
\usepackage{algorithmic}
\usepackage{graphicx}
\usepackage{textcomp}
\usepackage{xcolor}
\usepackage{algorithm}
\usepackage{algorithmic}
\usepackage{enumitem}
\usepackage{footnote}
\usepackage{multirow}
\usepackage[misc]{ifsym}

\ifCLASSOPTIONcompsoc
\usepackage[caption=false, font=normalsize, labelfont=sf, textfont=sf]{subfig}
\else
\usepackage[caption=false, font=footnotesize]{subfig}
\fi

\usepackage{array}

\def\BibTeX{{\rm B\kern-.05em{\sc i\kern-.025em b}\kern-.08em
    T\kern-.1667em\lower.7ex\hbox{E}\kern-.125emX}}

\begin{document}

\title{Natural Language Processing for Smart Healthcare}

\author{Binggui Zhou,
        Guanghua Yang\textsuperscript{\Letter},
        Zheng Shi,
        and Shaodan Ma\textsuperscript{\Letter}
\thanks{Binggui Zhou is with the School of Intelligent Systems Science and Engineering, Jinan University, Zhuhai 519070, China; and also with the State Key Laboratory of Internet of Things for Smart City and the Department of Electrical and Computer Engineering, University of Macau, Macao 999078, China.}
\thanks{ Guanghua Yang is with the School of Intelligent Systems Science and Engineering, Jinan University, Zhuhai 519070, China.}
\thanks{Zheng Shi is with the School of Intelligent Systems Science and Engineering, Jinan University, Zhuhai 519070, China; and also with the State Key Laboratory of Internet of Things for Smart City, University of Macau, Macao 999078, China.}
\thanks{
Shaodan Ma is with the State Key Laboratory of Internet of Things for Smart City and the Department of Electrical and Computer Engineering, University of Macau, Macao 999078, China.}
\thanks{
\textsuperscript{\Letter}Corresponding authors: Guanghua Yang (ghyang@jnu.edu.cn), Shaodan Ma (shaodanma@um.edu.mo).}
}

\markboth{IEEE Reviews in Biomedical Engineering}{Zhou \MakeLowercase{\textit{et al.}}: Natural Language Processing for Smart Healthcare}

\maketitle

\IEEEpubid{\begin{minipage}[t]{\textwidth}\ \\[12pt] \centering
  \copyright \ 2022 IEEE. Personal use of this material is permitted. Permission from IEEE must be obtained for all other uses, in any current or future media, including reprinting/republishing this material for advertising or promotional purposes, creating new collective works, for resale or redistribution to servers or lists, or reuse of any copyrighted component of this work in other works.
\end{minipage}} 

\begin{abstract}
Smart healthcare has achieved significant progress in recent years. Emerging artificial intelligence (AI) technologies enable various smart applications across various healthcare scenarios. As an essential technology powered by AI, natural language processing (NLP) plays a key role in smart healthcare due to its capability of analysing and understanding human language. In this work, we review existing studies that concern NLP for smart healthcare from the perspectives of technique and application. We first elaborate on different NLP approaches and the NLP pipeline for smart healthcare from the technical point of view. Then, in the context of smart healthcare applications employing NLP techniques, we introduce representative smart healthcare scenarios, including clinical practice, hospital management, personal care, public health, and drug development. We further discuss two specific medical issues, i.e., the coronavirus disease 2019 (COVID-19) pandemic and mental health, in which NLP-driven smart healthcare plays an important role. Finally, we discuss the limitations of current works and identify the directions for future works.

\end{abstract}


\section{Introduction}

\begin{figure*}[htbp]
\centering
\includegraphics[width=1\textwidth]{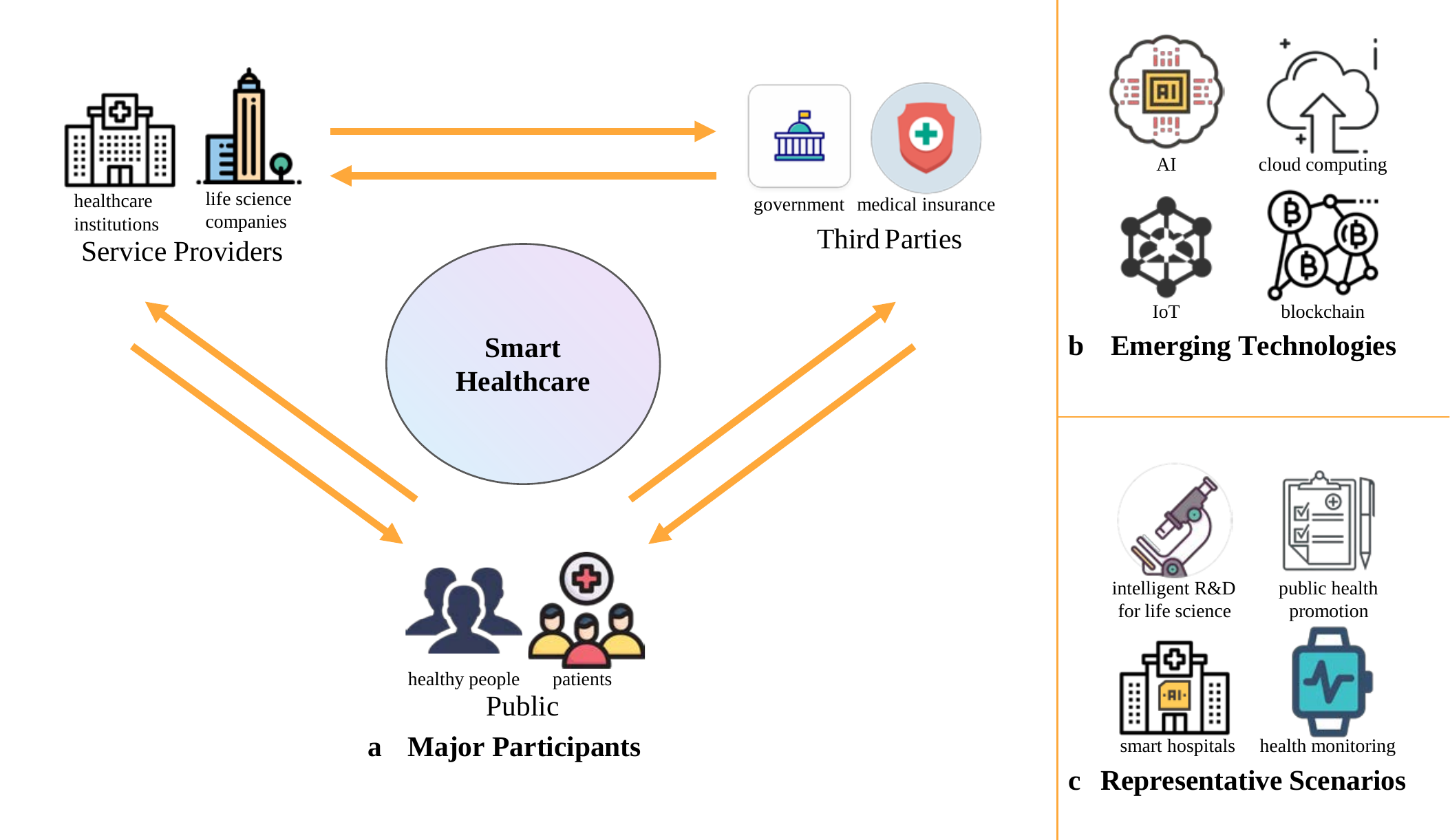}
\caption{\textbf{Smart healthcare}. \textbf{a}, major participants in smart healthcare include the public, healthcare service providers, and third-party healthcare participants. \textbf{b}, example emerging technologies enable smart healthcare applications include artificial intelligence, blockchain, cloud computing, the internet of things, and etc. \textbf{c}, representative smart healthcare scenarios include intelligent research and development for life science, public health promotion, smart hospitals, health monitoring, and etc.}
\label{smart_healthcare}
\end{figure*}

\IEEEPARstart SMART healthcare is a healthcare system that exploits emerging technologies, such as artificial intelligence (AI), blockchain, big data, cloud/edge computing, and the internet of things (IOT), for realizing various intelligent systems to connect healthcare participants and promote the quality of healthcare \cite{tian2019smart}. Major participants in smart healthcare can be classified into three categories, i.e., the public, healthcare service providers, and third-party healthcare participants. Related to the participants, representative smart healthcare scenarios include smart homes, smart hospitals, intelligent research and development for life science, health management, public health, rehabilitation therapy, and etc. Fig. \ref{smart_healthcare} shows the major participants, emerging technologies, and representative scenarios of smart healthcare. 

Natural language processing (NLP) is a subfield of computer science and artificial intelligence that is concerned with the automatic analysis, representation and understanding of human language \cite{young2018recent}. NLP has become a hot research area and has attracted widespread attention from many research communities in the past several years. As human language is a general form of data entry for intelligent systems, NLP enables machines to understand human language and interact with humans, making it essential to smart healthcare. 

The main manifestations of natural language are text and speech, where text encompasses text records, articles, book chapters, dictionaries, and so forth, while speech occurs in human-human and human-machine dialogues. NLP has been developed for several decades following the early origin of artificial intelligence in the 1950s. Approaches to conduct NLP are generally divided into three categories: rule-based approaches, statistical approaches, and deep learning-based approaches. From the 1950s to 1980s, NLP research mainly focused on rule-based approaches, which required expertise in both computer science and linguistics to design rules that fit human language. However, even well-designed rules are quite limited for covering human language due to its flexibility and complex patterns. Since the 1980s, statistical NLP systems have been designed by extracting features from corpora using statistical and machine learning algorithms and have gradually replaced rule-based NLP systems due to their superiority in performance and robustness. With the early application of the neural probabilistic language model \cite{bengio2003neural} and the rapid development of deep learning since 2013, neural NLP, by using neural networks and large corpora for automated feature learning, has dominated current research and achieved SOTA performance of many NLP tasks.

In smart healthcare, NLP is applied to process text data and is associated with human-machine/human-human communication. The text data can be classified into 2 categories: clinical text and other text data. Clinical text comes from all clinical scenarios and mainly comprises of unstructured text records from electronic health record (EHR) systems, including medical notes, diagnostic reports, electronic prescriptions, and etc. Other text data include all text that appears within other healthcare scenarios, e.g., surveys in population screening and articles for evidence-based reference. Communication is common in all smart healthcare scenarios, such as patient-provider communication in clinical inquiry and human-robot interaction in rehabilitation therapy, accompanied by applications such as machine translations and user interfaces for rehabilitation robots.

As well recognized, research on and applications of NLP for smart healthcare have received intensive attention in recent years. However, no study has offered a well-organized summary of existing works in a systematic way. In this paper, we first provide a systematic review of NLP for smart healthcare from both technical and application perspectives. After that, we discuss two specific medical issues, i.e., coronavirus disease 2019 (COVID-19) pandemic and mental health, in which NLP-driven smart healthcare plays an important role. Finally we discuss the limitations of existing works, identify the future directions of applying NLP to smart healthcare, and close the review with some conclusions.

\section{NLP for smart healthcare from technical perspective}

NLP has been undergoing continuous development since the 1950s. Studies on NLP for smart healthcare have also been conducted for decades and have attracted increased attention in recent years with the advancement of artificial intelligence and general NLP. To connect existing works from technical perspective, in this section, we first introduce the three kinds of NLP approaches and their representative algorithms, and then introduce the NLP pipeline for smart healthcare to show how NLP techniques are used in real smart healthcare applications.

\subsection{Comparisons of different NLP approaches}

The mainstream NLP approaches can be classified into three categories, i.e., rule-based NLP, statistical NLP and neural NLP, which have different characteristics. Below, we discuss the advantage and disadvantages of the three categories and introduce the representative algorithms of them.

Rule-based NLP approaches, e.g., pattern matching \cite{crim2005automatically} and parsing \cite{vilares2008extraction}, could be quite accurate in specific cases if dedicated studies by experts are conducted. In addition, rule-based NLP approaches are easy to interpret and understand. However, rules are normally too limited to cover all cases considering the flexibility and complex patterns of human language. In addition, rule-based NLP requires expertise in both computer science and linguistics to design appropriate rules to fit human language, hindering it from large-scale applications. Currently, rule-based approaches have been widely considered obsolete by academia \cite{chiticariu2013rulebased}, and are occasionally used for better preprocessing nowadays \cite{kang2012using}.

In general, statistical NLP is superior to rule-based NLP in performance and robustness. However, it also requires domain expertise to create handcrafted features, and is therefore limited to taking full advantage of available data and providing enough accuracy in complex applications. Although statistical NLP requires intensive feature engineering, it is this direct feature design that makes it transparent and interpretable as rule-based NLP. In addition, statistical NLP does not rely on large-scale datasets or large amounts of computational power, and thus is much more efficient than neural NLP. Furthermore, representative statistical NLP models, such as bag-of-words \cite{weng2017medical}, TF-IDF \cite{dessi2020tfidf,ozyegen2021wordlevel}, and n-gram \cite{rahimian2019significant,yazdani2019words,yip2010concept}, have different characteristics. Bag-of-words is easy to implement, but it only considers the frequencies of words in a sentence, which neglects the importance and sequential order of these words. Through the inverse document frequency, TF-IDF improves the measurement of a word's importance, but still does not take sequential order information into consideration. N-gram considers $n-1$ words before a word, which makes it more accurate than bag-of-words but with higher computational complexity (increases exponentially with $n$). It is worth mentioning that despite the dominance of deep learning in recent years, statistical NLP is still active in many healthcare studies and applications.

Recent years have witnessed the success of neural NLP, who has shown better performance than both rule-based NLP and statistical NLP in applications with abundant available data. However, neural NLP is often blamed for low interpretability and dependence on expensive computing platforms. It is also worth noting that, compared with rule-based NLP and statistical NLP, neural NLP usually fails to achieve satisfactory performance if limited data is available. Among neural NLP models, recurrent neural network (RNN)-based models, especially long short-term memory (LSTM) \cite{beeksma2019predicting,jagannatha2016bidirectional,liu2016augmented}-based models and gated recurrent unit (GRU)-based models \cite{jagannatha2016bidirectional,zhao2018leveraging}, are more natural for processing sequential data such as text and speech. They have the ability to remember historical information of the inputs, but suffer from gradient vanishing/explosion, training issues and short-term memories. Convolutional neural networks (CNN)-based models \cite{hughes2017medical,li2019intelligent}, combining with word embeddings, also show good performance in some tasks due to their ability in learning local features and high computational efficiency which enables deep network architectures. Recently, graph neural network (GNN)-based models have been applied to NLP-driven smart healthcare by incorporating knowledge from graph-structured ontology/entities \cite{li2020graph,sun2021disease,wu2021leveraging}. When graphs are large in scale or complex, GNN-based models are difficult and costly to implement and train. Generally speaking, RNNs, CNNs, and GNNs are all limited in tackling long-term dependencies in sequences. Through the self-attention mechanism, Transformer-based models \cite{mayer2020transformerbased,zhang2020timeaware} are very efficient in processing long sequences and support parallel training, but are lack of ability in learning local features and position information. We have witnessed many combinations of the aforementioned models for better feature extraction performance, including CNN-LSTM networks \cite{tokala2018deep}, RNN-Attention networks \cite{choi2017retain,chu2018using}, memory networks (MM) \cite{chakraborty2020explicitblurred,song2021local}, graph convolutional networks (GCN) \cite{yoon2019information}, CNN-LSTM-Attention networks \cite{cai2017cnnlstm,tang2019entity}, graph convolutional attention networks (GCAN) \cite{choi2020learning,wang2020documentlevel}, etc. In addition, to further leverage large unlabelled corpora, pretraining, a very effective method, has been widely exploited to obtain non-contextual or contextual embeddings \cite{qiu2020pretrained}. Word2vec \cite{beam2019clinical,cai2018medical,kholghi2016benefits,zhang2019biowordvec}, and GloVe (Global Vectors) \cite{beam2019clinical,dubois2018effective}, as representative algorithms of non-contextual embeddings, provide distributed dense vectors as word embeddings, and outperform statistical algorithms such as bag-of-words and n-gram. The non-contextual embedding for a word is static and does not dynamically change as its context changes \cite{qiu2020pretrained}. Based on the Transformer architecture, contextual embeddings, e.g., ELMo (Embeddings from
Language Models) \cite{jin2019probing}, BERT (Bidirectional Encoder Representations from Transformers) \cite{lee2020biobert,huang2020clinicalbert,rasmy2021medbert,li2020behrt}, and GPT (Generative Pre-Training) \cite{schneider2021gpt2}, are developed to embed dynamic contextual information into word embeddings, achieving outstanding performance than other word embedding algorithms. It should be noted that these models are typically huge and expensive to pre-train, which somehow constraints their broad application in healthcare.

The comparisons of different NLP approaches and representative algorithms are shown in Table \ref{features}.

\subsection{NLP pipeline for smart healthcare}

\begin{figure*}[htbp]
\centering
\includegraphics[width=1\textwidth]{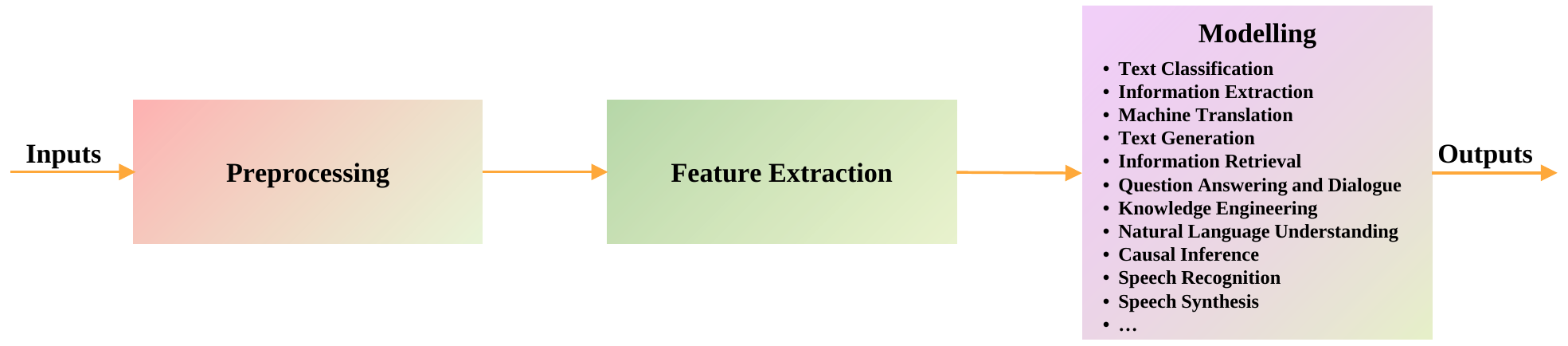}
\caption{\textbf{The NLP pipeline for smart healthcare}. There are three parts in an NLP pipeline for smart healthcare, i.e., preprocessing, feature extraction, and modelling. NLP takes text or speech as the input, followed by preprocessing to facilitate feature extraction and modelling. Features can be extracted with various methods and models. Models for specific NLP tasks are finally built with the extracted features to yield the outputs.}
\label{nlp_pipelines}
\end{figure*}

As shown in Fig. \ref{nlp_pipelines}, there are three parts in an NLP pipeline for smart healthcare, i.e., preprocessing, feature extraction, and modelling. An NLP pipeline takes text or speech as illustrated before as the input. After that, preprocessing is conducted considering various inputs and their qualities to facilitate feature extraction and modelling. As the most important step, feature extraction is essential to NLP, which undoubtedly explains the attention it has received from researchers. Finally, models for specific NLP tasks are built with the extracted features to yield the outputs accordingly.

\subsubsection{Preprocessing}
Preprocessing, including the procedures of tokenization, stemming, lemmatization, stopword removal, and etc., makes natural language normalized, machine-readable, and easy for postprocessing. Text preprocessing mostly paves the way for feature extraction and modelling, since many NLP tasks require normalized text input to guarantee accuracy and efficiency due to significant challenges coming from the flexibility of natural languages and the wide variety of morphological variants of medical terms in medical text \cite{akkasi2016chemtok,dai2015enhancing,liu2012biolemmatizer}. However, with the development of neural NLP, some text preprocessing procedures have become unnecessary and may even cause problems. For example, removing stopwords may lead to the loss of informative context information when using the BERT pre-trained model \cite{qiao2019understanding}. As the preprocessing of speech, such as denoising, is typically regarded as a problem in signal processing, we do not discuss it in detail here.

\subsubsection{Feature extraction}
Apart from the increase in accessible digital data and the advances in computing platforms such as graphics processing units, the development of NLP is largely attributed to the improvement in feature design or feature extraction methods. Both rule-based approaches and statistical approaches require expertise for rule design \cite{crim2005automatically,vilares2008extraction} or feature engineering \cite{weng2017medical,dessi2020tfidf,ozyegen2021wordlevel,rahimian2019significant,yazdani2019words,yip2010concept}. For neural NLP, automated feature extraction via varieties of neural networks \cite{beeksma2019predicting,jagannatha2016bidirectional,liu2016augmented,zhao2018leveraging,hughes2017medical,li2019intelligent,li2020graph,sun2021disease,wu2021leveraging,mayer2020transformerbased,zhang2020timeaware,tokala2018deep,choi2017retain,chu2018using,chakraborty2020explicitblurred,song2021local,yoon2019information,cai2017cnnlstm,tang2019entity,choi2020learning,wang2020documentlevel} have greatly improved the efficiency of data utilization and feature extraction. Automated feature engineering can be conducted directly according to the downstream tasks using supervised learning, unsupervised learning or reinforcement learning. In addition, pretraining is also widely used in NLP to automatically extract features from large unlabelled corpora via self-supervised learning in a generative, contrastive or generative-contrastive manner \cite{liu2021selfsupervised} before the downstream tasks begin. The extracted features, known as contextual or non-contextual embeddings, may encompass features such as lexical meanings, syntactic features, semantic features, and even pragmatics, which contribute to downstream tasks \cite{qiu2020pretrained}.

\begin{table*}[htbp]
\centering
\caption{Comparisons of different NLP approaches and representative algorithms.}
\label{features}
\resizebox{\textwidth}{!}{%
\begin{tabular}{p{2cm}|p{2cm}|p{4.2cm}|p{7cm}}
\hline
\textbf{NLP   Approach} & \textbf{Feature Extraction Method} & \textbf{Advantages and Disadvantages} & \textbf{Representative Algorithms} \\ \hline
\textbf{Rule-based NLP} & rule design & \begin{tabular}[c]{@{}l@{}}\textbf{advantages}:\\    -  could be quite accurate in specific\\ cases;\\   -   easy to interpret and understand\\      \\      \textbf{disadvantages}:\\    -  rules are too limited to cover all\\ cases considering the flexibility and\\   complex patterns of human language;\\  -    require expertise in both computer\\ and linguistics to fit human language\end{tabular} & pattern matching \cite{crim2005automatically} and   parsing \cite{vilares2008extraction} \\ \hline
\textbf{Statistical NLP} & hand-crafted feature engineering & \begin{tabular}[c]{@{}l@{}}\textbf{advantages}:\\  -    superior to rule-based NLP in\\ performance and robustness;\\  -    good interpretability\\      \\      \textbf{disadvantages}:\\  -    require domain expertise to create\\ handcrafted features;\\ -     limited to taking full advantage of\\ available data and providing enough\\   accuracy in complex applications\end{tabular} & \begin{tabular}[c]{@{}l@{}}bag-of-words \cite{weng2017medical}:\\  -   easy to implement;\\      - neglects the importance and sequential order of words\\      \\      TF-IDF \cite{dessi2020tfidf,ozyegen2021wordlevel}:\\    -  improves the measurement of a word's importance;\\   -   does not take sequential order information into consideration\\      \\      n-gram \cite{rahimian2019significant,yazdani2019words,yip2010concept}:\\     -  more accurate than bag-of-words;\\ - high computational complexity (increasing exponentially with $n$)\end{tabular} \\ \hline
\textbf{Neural NLP} & automated feature extraction & \begin{tabular}[c]{@{}l@{}}\textbf{advantages}:\\    -  better performance than both\\ rule-based NLP and statistical NLP\\ in  applications with abundant\\ available data\\      \\      \textbf{disadvantages}:\\    -  low interpretability;\\      - dependence on expensive computing\\ platforms;\\    -  usually fail to achieve satisfactory\\ performance if limited data is\\ available\end{tabular} & \begin{tabular}[c]{@{}l@{}} 1) RNN-based models (e.g., LSTM \cite{beeksma2019predicting,jagannatha2016bidirectional,liu2016augmented} and GRUs \cite{jagannatha2016bidirectional,zhao2018leveraging}):\\      - more natural for processing text and speech input;\\  - capable of remembering historical information of the inputs;\\   -   suffer from gradient vanishing/explosion, training issues and\\ short-term   memories\\      \\      2) CNN-based models \cite{hughes2017medical,li2019intelligent}:\\     - able to learn local features;\\ - high computational efficiency;\\    -  limited in tackling long-term dependencies in sequences\\      \\      3) GNN-based models \cite{li2020graph,sun2021disease,wu2021leveraging}:\\ -     efficient in incorporating knowledge from graph-structured\\   ontology/entities;\\ -     limited in tackling long-term dependencies in sequences;\\      - difficult and costly to implement and train with large-scale\\ or very complex   graphs\\ \\ 4) Transformer-based models \cite{mayer2020transformerbased,zhang2020timeaware}:\\   -   efficient in processing long sequences and parallel training;\\  -    lack of ability in learning local features and position information\\      \\      5) combinations: CNN-LSTM \cite{tokala2018deep},   RNN-Attention \cite{choi2017retain,chu2018using},\\   MN \cite{chakraborty2020explicitblurred,song2021local},   GCN \cite{yoon2019information},   CNN-LSTM-Attention \cite{cai2017cnnlstm,tang2019entity},\\ and   GCAN \cite{choi2020learning,wang2020documentlevel}, etc.\\ \\ 6) non-contextual embedding-oriented pre-trained models\\ (word2vec \cite{beam2019clinical,cai2018medical,kholghi2016benefits,zhang2019biowordvec}, GloVe \cite{beam2019clinical,dubois2018effective}):\\     - outperform statistical algorithms;\\   -   the non-contextual embedding for a word is static and will not\\ dynamically   change as its context change\\      \\      7) contextual embedding-oriented pre-trained models (ELMo \cite{jin2019probing},\\   BERT \cite{lee2020biobert,huang2020clinicalbert,rasmy2021medbert,li2020behrt}, GPT \cite{schneider2021gpt2}):\\  - able to embed dynamic contextual information into word embeddings;\\ - outstanding performance than other word embedding algorithms;\\ - typically huge and expensive to pre-train\end{tabular} \\ \hline
\end{tabular}
}
\end{table*}

\subsubsection{Modelling}
For various smart healthcare applications, different models should be built to accomplish various NLP tasks, such as text classification, information extraction, and natural language understanding. The extracted feature can be directly processed by classifiers and regressors to yield outputs for simple tasks, e.g., medical text classification \cite{hughes2017medical,wang2019clinical}, while further steps are required to complete complex tasks. In the following subsections, we first introduce several text input-based NLP tasks according to their complexity. At the end of this section, we will introduce two speech-specific tasks, i.e., speech recognition and speech synthesis.

\textbf{Information extraction}. Information extraction (IE), a.k.a. text mining, enables harvesting information from text inputs, and plays an important role in text analysis. Works related to information extraction in smart healthcare focus on the extraction of diseases, drugs, events (mainly including temporal expressions, spatial expressions and participant information) through name entity recognition \cite{hassanpour2016information,perera2020named}, relation extraction \cite{perera2020named,thillaisundaram2019biomedical,zitnik2015sievebased}, and event extraction \cite{jindal2013extraction} from medical text, including unstructured text in EHRs, articles, etc.

\textbf{Machine translation}. Machine translation (MT) aims to automatically translate text from one language to another \cite{garg2018machine}. Currently, healthcare resources in various languages are becoming easily accessible as technologies evolve, and they are all of great value in modern medical practice. Machine translation therefore has drawn growing attention for building better (multilingual) translation systems and further leveraging multilingual healthcare resources for other applications, either to provide more accurate translations \cite{berard2020multilingual,kirchhoff2011application} or to require less time \cite{kirchhoff2011application} than human translations.

\textbf{Text generation}. Text generation (TG) automatically generates text with given inputs while pursuing the goal of appearing indistinguishable from human-written text. Specifically, there are 3 kinds of inputs and corresponding subtasks in smart healthcare: text inputs (e.g., routine reports) associated with text summarization \cite{afzal2020clinical,lopez2019automatic,manas2021knowledgeinfused}, question generation \cite{pistol2018meditest,shen2019generation,wang2008automatic}, dialogue generation \cite{li2021semisupervised,lin2020graphevolving,yang2020generation}, and etc.; data inputs (e.g., neonatal intensive care data) connected with data-to-text \cite{pauws2018making}; and image inputs (e.g., medical images) related to image captioning \cite{kougia2019survey,xiong2019reinforced}, visual question answering (VQA) \cite{he2021visual,zhan2020medical,afrae2019new}, and etc. Note that for data-to-text and image-to-text generation, a combination of NLP with data analysis or computer vision is generally required, respectively.

\textbf{Information retrieval}. Information retrieval (IR) obtains materials that meet the query requirements from numerous documents, and is a core of search engines for all applications. To ease the retrieval process \cite{wu2018semehr,gobeill2015deep}, improve the relevance and diversity of the retrieval \cite{montazeralghaem2020relevance,xu2018improve,urbain2009passage} or reduce the query time \cite{mohan2018fast}, current works aim to develop fast and efficient information retrieval methods to obtain useful retrieval from a large collection of data sources, ranging from internal health information system (HIS) systems and other digital documents to online resources.

\textbf{Question answering and dialogue systems}. Question answering (QA) involves automatically providing answers to questions raised by humans in a natural language. Question answering requires the machine to understand natural language and infer the answers, making it highly dependent on natural language understanding and information retrieval. To date, QA systems for healthcare have developed from information retrieval based QA systems \cite{hristovski2015biomedical,vinod2021multilingual,zahid2018cliniqa,wang2020texttosql} and knowledge-based QA systems \cite{jiang2021research,liu2017knowledgebased,demner-fushman2007answering,terol2007knowledge} to hybrid QA systems \cite{liu2018tknow,mutabazi2021review}. Compared to question answering, dialogue is also presented in an interactive manner between humans and machines. Common dialogue systems in smart healthcare include task-oriented dialogue systems \cite{shim2021building,wei2018taskoriented,xu2019endtoend}, and non-task-oriented (a.k.a. chat-oriented) \cite{kawata2017lifestyle} dialogue systems, which assume different functions in various applications.

\textbf{Knowledge engineering}. Knowledge engineering (KE) is a field within artificial intelligence that tries to construct and use knowledge-based systems \cite{studer1998knowledge}. It does not refer to a pure NLP technique, but receives much attention in NLP for smart healthcare since medical text is one of the major sources for knowledge engineering. Within knowledge engineering, knowledge acquisition and knowledge representation are coupling with information extraction, aiming at the acquisition and representation of medical knowledge in a certain way, e.g., knowledge graphs \cite{goodwin2013automatic,rossanez2020kgen,rotmensch2017learning}. Besides, knowledge engineering also concerns building knowledge-based systems to exploit existing knowledge, such as knowledge-based question answering (KBQA) systems \cite{jiang2021research,liu2017knowledgebased,demner-fushman2007answering,terol2007knowledge}, knowledge-based information retrieval systems \cite{wang2017semantically}, text generation systems \cite{pan2020medwriter,shen2019generation}, etc.

\textbf{Natural language understanding}. Natural language understanding (NLU) focuses on machines’ comprehension of human language in the form of unstructured text or speech. Many of the aforementioned tasks, e.g., question answering, information retrieval, require NLU to fully understand the input queries. The difficulties of natural language understanding come from the diversity, ambiguity, and potential dependence of natural language, making slow progress in natural language understanding compared with other NLP techniques. After years of development in both general areas and smart healthcare, the mainstream route of NLU is still to use various methods to conduct slot filling and intent detection \cite{stoica2021intent,neuraz2018natural,zhang2017bringing}. NLU is the core of multiple intelligent agents, assuming a role in understanding human intentions during human-machine interactions \cite{stoica2021intent,giachos2017exploring,thomason2019improving}, medical queries \cite{neuraz2018natural,zhang2017bringing}, etc.

\textbf{Causal inference}. Generally, causal inference is a discipline concerning the determination of actual effects of specific things, events or phenomena. Causal inference in NLP has long received insufficient attention since the goal of classical NLP applications is simply to make accurate predictions with all available statistical correlations regardless of the underlying causal relationship \cite{feder2021causal}. Recently, with growing concerns about uninterpretable black box models, the importance of causal inference has gradually been recognized by NLP researchers, especially in the area of healthcare. Specifically, recent advances of causal inference in NLP for smart healthcare have been made in uncovering causality from medical text \cite{zeng2021uncovering,doan2019extracting,nordon2019building} and realizing reliable NLP-driven applications with discovered causal effects \cite{zeng2021uncovering,doan2019extracting,nordon2019building}.

\textbf{Speech recognition and speech synthesis}. Speech recognition (SR) aims to convert human speech into text information. Contrary to speech recognition, speech synthesis, a.k.a. text-to-speech (TTS), is concerned with representing text information with speech. Basically, SR-oriented and SS-oriented studies attempt to build automatic computer systems for interconversion between speech and text in the area of smart healthcare, making human-machine interaction as natural and flexible as human-human interaction \cite{stiefelhagen2004natural}. For speech recognition, these efforts encompass the improvement in acoustic modelling \cite{finley2018semisupervised,sas2011optimal}, language modelling \cite{paulett2009improving}, and the whole system pipeline \cite{chiu2018speech,edwards2017medical} to enhance recognition accuracy. For speech synthesis, recent advancements have been made in investigating and making synthesized speech natural \cite{he2020doptacotron,sugiura2014nonmonologue}, intelligible \cite{akbari2019reconstructing,anumanchipalli2019speech,herff2019generating,jreige2009vocalid,marge2022spoken} and expressive \cite{james2020empathetic,li2009expressive,roehling2006expressive}, which will help stimulate the enthusiasm of human-machine interaction \cite{kuhne2020human}.

\section{Applications of NLP for smart healthcare}

\begin{figure}[htbp]
\centering
\includegraphics[width=0.5\textwidth]{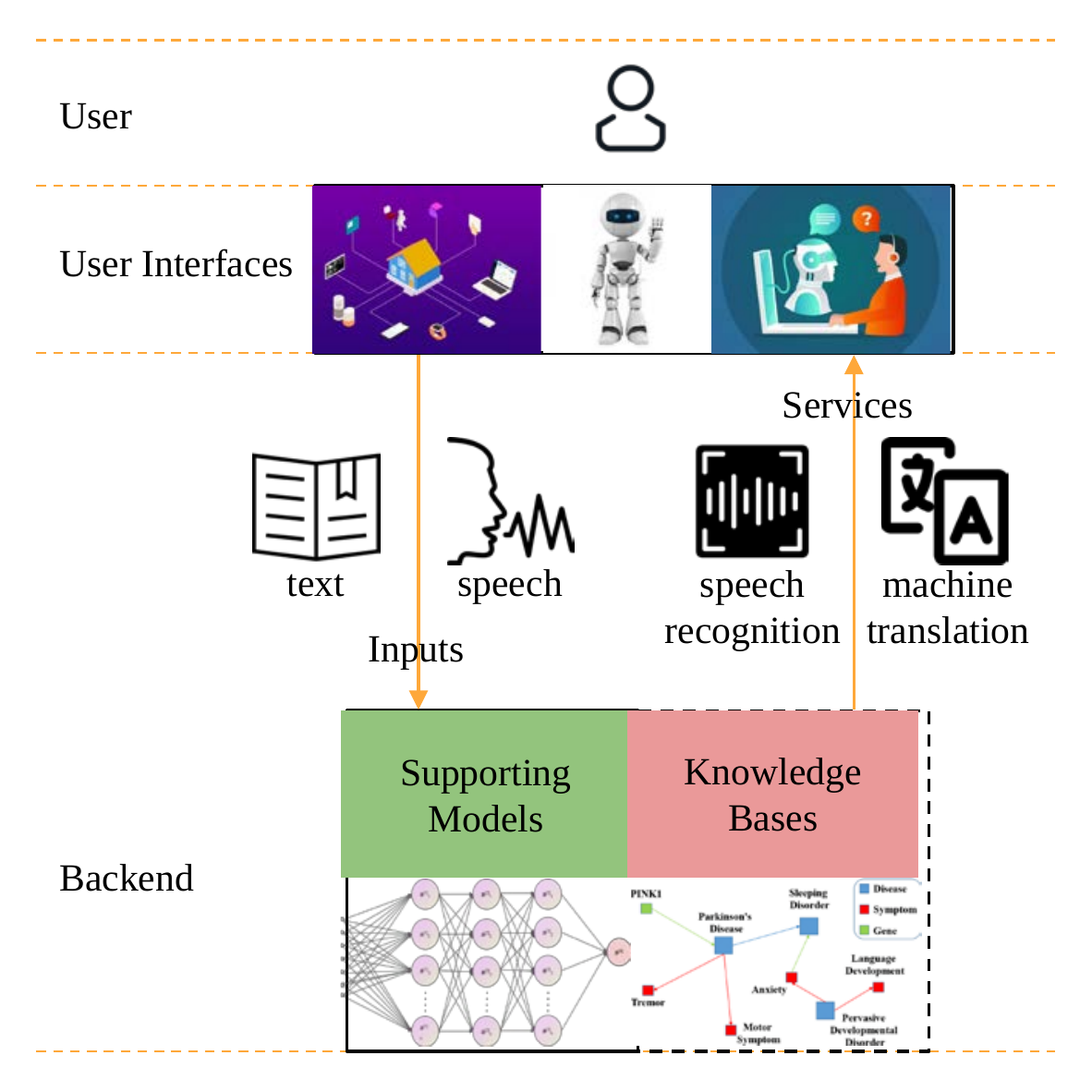}
\caption{\textbf{Basic architecture of NLP-driven applications}. A typical NLP-driven application is composed of user interface and backend, where the UI takes inputs from the user and feedback the results to the user, and the backend processes these inputs with the NLP models with or without the knowledge bases according to the specific task type.}
\label{nlp_driven_apps}
\end{figure}

NLP has been widely applied in smart healthcare and has brought dramatic improvements in many applications. As shown in Fig. \ref{nlp_driven_apps}, a typical NLP-driven application is composed of two parts: user interface (UI) and backend. The user provides text or speech input to the backend through the UI, and then, the backend processes these inputs with the NLP models and feeds the results back to the user by providing specific services through the UI. Knowledge bases are also required at the backend for applications that essentially rely on knowledge, for example, the aforementioned KBQA systems. The NLP techniques described in the previous section play a key role in both UI and backend. 

The UI enables information exchange between users and intelligent systems through speech, text, etc. Easily accessible UIs are critical for enhancing the experience of using intelligent systems and realizing smart healthcare. Such user interfaces can be implemented by using NLP techniques, especially speech recognition and natural language understanding.

According to their application scenarios, smart healthcare applications employing NLP techniques can be classified into 5 major categories, i.e., clinical practice, hospital management, personal care, public health, and drug development. A summary of the applications and related NLP techniques is presented in Table \ref{applications}. Below we introduce the five categories in detail.

\subsection{Clinical practice}
\textbf{Clinical communication and data collection}. Clinical data, including but not limited to demographics, medical history, comorbidities, medical notes, physical examination notes, electronic recordings from medical devices, and clinical laboratory testing data and medical images \cite{jiang2017artificial}, are the most important data for diagnosis, treatment and even further retrospection. Patient-provider communication is an important way to obtain first-hand clinical data. When necessary, machine translation may assist doctors in communicating with patients who speak different languages or have low literacy and limited levels of health education \cite{dew2018development,randhawa2013using}. Meanwhile, free text notes can be taken through speech recognition \cite{goss2019clinician,saxena2018provider,zhao2009speechrecognition}, which can significantly reduce medical staff's time on labour-intensive clinical documentation.

\textbf{Clinical decision support}. Clinical decision support (CDS) systems can provide physicians with diagnosis and treatment suggestions, which play an increasingly important role in clinical medicine with the surge of clinical cases and growing concerns regarding public healthcare. With the development of question answering systems, clinical decision support based on question answering \cite{goodwin2016medical,zahid2018cliniqa,xu2021external} has emerged and become common, as it is closer to traditional patient-provider communication. NLP techniques have shown great ability to build clinical decision support systems by extracting various useful information for making diagnosis and therapeutic decisions, such as family history information \cite{shi2019family}, entities and relations \cite{gupta2018automatic,yang2019information}, treatment and prognosis data \cite{zheng2018automated}, clinical data concepts and features \cite{liang2019evaluation}, and even causal relations \cite{doan2019extracting,nordon2019building}. In addition, to ensure quality control and future quality improvement, NLP can also contribute to the assessment of clinical procedures \cite{harkema2011developing,mehrotra2012applying}, warning of potentially harmful adverse drug events (ADEs) \cite{wunnava2019adverse}, disease symptoms \cite{jackson2017natural,luo2021early}, and outcome-related causal effects \cite{wang2021inferbert}. Finally, NLP is also powerful in enhancing the interpretability and reliability of clinical decision support systems for practical deployment, by providing supporting evidence for diagnosis or treatment decisions in an evidence-based fashion \cite{zeng2021uncovering,wang2019transparent,lee2019information,patrick2004evidencebased,ford1999information}.

\subsection{Hospital management}
\textbf{Medical resource allocation}. Due to limited medical resources, including hospital spaces, personnel, and materials, efficient resource allocation is critical in hospitals and other medical facilities. By building patient triage systems, medical resources can attend to critical cases with priority and enhance medical resource allocation effectiveness and efficiency \cite{sterling2019prediction,tahayori2021advanced}. Virtual assistants \cite{sezgin2020readiness,spanig2019virtual,gandhi2019intellidoctor}, hospital automation systems \cite{agustin2019voice,ismail2020development} and collaborative robots \cite{grasse2021speech,holland2021service} with voice control can further reduce the burden on medical staff, thereby improving hospital management efficiency. There are also some interesting works that have explored the prediction of patient readmission to rearrange medical resources/interventions and reduce the readmission rate \cite{lineback2021prediction,rajkomar2018scalable,rumshisky2016predicting}. In addition, by leveraging text generation techniques, part of text writing in healthcare, especially routine reports, can be taken over by machines, freeing medical staff from many administrative duties and making them available for direct patient care \cite{alfarghaly2021automated,pauws2018making}.

\textbf{Data management}. To manage large volumes of medical documentation, text classification, information extraction and text summarization can be used to generate category labels, informative keywords and simplified summaries \cite{hughes2017medical,wang2019clinical,lopez2019automatic,manas2021knowledgeinfused,chintagunta2021medically} for management, while information retrieval systems, especially those systems based on semantic search \cite{wu2018semehr,ibrahim2013ontologydriven} and question answering \cite{gobeill2015deep}, can be used in healthcare information systems to ease the retrieval process.

\textbf{Service quality control}. Sentiment analysis with patient experience feedback will help hospitals improve their service quality and patient experience. Such analysis required substantial personnel resources in the past, while NLP makes this work easier and greatly improves the efficiency of sentiment analysis \cite{khanbhai2021applying,nawab2020natural,doing-harris2017understanding}.

\subsection{Personal care}

\textbf{Personal health assistants}. Personal health assistants enable people to easily access useful medical information and healthcare services without visiting the healthcare institutions. Personal health assistants may incorporate several subsystems, such as medical information access systems \cite{rodger2013pregnant} and remote healthcare systems \cite{zhou2018new}, for various purposes. 

\textbf{Assisting elderly individuals and disabled individuals}. NLP techniques can help elderly individuals and disabled individuals to greatly enhance their quality of life and social integration. Voice-controlled home automation systems and robots may assist the elderly and the disabled in their daily lives \cite{rani2017voice}, while robots (especially androids and other robots that communicate with people) can even encourage and accompany them through social interactions \cite{tapus2007socially,mavridis2015review}. In addition, NLP techniques are also of great value for providing essential aids to people with various disabilities, e.g., speech impairments \cite{green2021automatic,hux2017comprehension,hux2020effect,cassidy2016expressive,repova2020texttospeech,jreige2009vocalid}, hearing loss \cite{lyall2016smartphone}, dyslexia \cite{nittrouer2018speech}, or neurological disorders \cite{akbari2019reconstructing,anumanchipalli2019speech,herff2019generating}.

\subsection{Public health}
\textbf{Health knowledge popularization and medical education}.
Health knowledge popularization and medical education are essential public health interventions since they can improve people's health literacy and help them develop healthy living habits. Through knowledge engineering, accurate and complete medical knowledge bases can be established to promote the popularization of medical knowledge among the population \cite{goodwin2013automatic,rossanez2020kgen,rotmensch2017learning,riano2019ten,demner-fushman2007answering,jiang2021research,liu2017knowledgebased,terol2007knowledge,wang2017semantically}. Specifically, people can easily access medical knowledge through question answering systems \cite{bao2020hhh,xie2018mobilebased}, information retrieval systems \cite{xu2018improve,mohan2018fast}, and machine translation systems \cite{peters2016translating,renato2018machine,tian2019smart,randhawa2013using}, facilitating the popularization and education of medical knowledge. In addition, text generation techniques, such as question generation and text summarization, can also be used in medical education to generate medical case-based questions \cite{leo2019ontologybased} and construct simplified summaries \cite{afzal2020clinical}.

\textbf{Population screening}. In addition to the health knowledge popularization, population screening, which refers to the process of assessing the prevalence of a disease or condition in a population or subgroup, is also an important intervention for delivering public health. The population screening starts with identifying target populations, followed by the screening test. After that, further actions such as further tests, advice, or treatment can be taken considering the screening results \cite{nhs_2013}. NLP can play two main roles in population screening. First, NLP helps identify populations with higher health risk factors, which may improve the efficiency of population screening \cite{m2015state}. Second, NLP can also assist in the analysis of healthcare questionnaires and surveys \cite{georgiou2015extracting}, especially for open-ended questions.

\subsection{Drug development}

\textbf{Drug discovery}. NLP helps construct textual representations of biochemical entities for mapping the interactions between diseases, drugs/chemical compounds, and biomacromolecules (e.g., genes, proteins); predicting molecular properties; and designing novel molecules. Readers are referred to the comprehensive review by Öztürk et al \cite{ozturk2020exploring} for a deeper understanding of NLP methodologies for drug discovery.

\textbf{Preclinical research}. NLP techniques, especially information extraction, are also able to identify the relations between chemical structures and biological activity \cite{lake2019artificial} and further help researchers search for potentially effective chemical compounds, i.e., virtual screening \cite{pham2021deep,schwartz2013smifp}, in a huge chemical space. In addition, they are also applied in the prediction of adverse drug reactions, including side effect prediction \cite{zhang2021prediction}, toxicity prediction \cite{bouhedjar2020natural,jeon2019fp2vec}, and etc., in preclinical research.

\textbf{Clinical research}. Across the clinical research stage, NLP may enable efficient clinical trial design \cite{nordon2019building}, patient recruitment \cite{chen2019clinical,harrer2019artificial,tissot2020natural}, clinical trial analytics \cite{chen2020trends}, and etc. 

\textbf{Drug review and safety monitoring}. Recently, the FDA and other institutions have reported being interested in using NLP for adverse drug event discovery and drug safety monitoring \cite{ventola2018big,wang2009active,liu2019drug}, showing the full range of NLP's key role in drug development.

\begin{table*}[htbp]
\centering
\caption{Applications driven by NLP in all Smart Healthcare Scenarios.}
\label{applications}
\resizebox{\textwidth}{!}{%
\begin{tabular}{p{2.5cm}|p{3cm}|p{9cm}|p{4cm}}
\hline
\textbf{Category} &
  \textbf{Sub-Category} &
  \textbf{Representative Applications} &
  \textbf{Related Techniques} \\ \hline
\multirow{6}{*}{Clinical Practice} &
  \multirow{2}{3cm}{clinical communication and data collection} &
  patient-provider   communication \cite{dew2018development,randhawa2013using} &
  machine translation \\ \cline{3-4} 
 &
   &
  clinical   documentation \cite{goss2019clinician,saxena2018provider,zhao2009speechrecognition} &
  speech recognition \\ \cline{2-4} 
 &
  \multirow{4}{*}{clinical decision   support} &
  build QA-based clinical decision support   systems \cite{goodwin2016medical,zahid2018cliniqa,xu2021external} &
  information extraction \\ \cline{3-4} 
 &
   &
  build clinical decision   support systems with extracted information: family history   information \cite{shi2019family}, entities and   relations \cite{gupta2018automatic,yang2019information}, treatment and   prognosis data \cite{zheng2018automated}, clinical data concepts and   features \cite{liang2019evaluation}, causal   relations \cite{doan2019extracting,nordon2019building} &
  question answering \\ \cline{3-4} 
 &
   &
  healthcare quality control:   assess clinical procedures \cite{harkema2011developing, mehrotra2012applying},   warning of ADE \cite{wunnava2019adverse}, disease   symptoms \cite{jackson2017natural,luo2021early}, and outcome-related causal   effects \cite{wang2021inferbert} &
  information extraction, causal inference \\ \cline{3-4} 
 &
   &
  provide supporting evidence   for decisions under evidence-based   fashion \cite{zeng2021uncovering,wang2019transparent,lee2019information,patrick2004evidencebased,ford1999information} &
  information retrieval, causal inference \\ \hline
\multirow{7}{*}{Hospital   Management} &
  \multirow{4}{*}{medical resource allocation} &
  patient   triage \cite{sterling2019prediction,tahayori2021advanced} &
  information extraction \\ \cline{3-4} 
 &
   &
  enable   users to communicate and control intelligent systems through virtual   assistants \cite{sezgin2020readiness,spanig2019virtual,gandhi2019intellidoctor},   hospital automation systems \cite{agustin2019voice,ismail2020development} and   collaborative robots \cite{grasse2021speech,holland2021service} &
  speech recognition, natural language understanding\\ \cline{3-4} 
 &
   &
  predict and reduce   readmission   rate \cite{lineback2021prediction,rajkomar2018scalable,rumshisky2016predicting} &
  information extraction \\ \cline{3-4} 
 &
   &
  free medical staff from   routine text writing \cite{alfarghaly2021automated,pauws2018making} &
  information extraction \\ \cline{2-4} 
 &
  \multirow{2}{*}{data management} &
  manage clinical   documents \cite{hughes2017medical,wang2019clinical,lopez2019automatic,manas2021knowledgeinfused,chintagunta2021medically} &
  text generation \\ \cline{3-4} 
 &
   &
  ease the HIS retrieval   process based on semantic search \cite{wu2018semehr,ibrahim2013ontologydriven}   and question answering \cite{gobeill2015deep} &
  text classification, text summarization, information   extraction \\ \cline{2-4} 
 &
  service quality control &
  improve service quality and patient   experience \cite{khanbhai2021applying,nawab2020natural,doing-harris2017understanding} &
  information retrieval, question answering \\ \hline
\multirow{5}{*}{Personal Care} &
  \multirow{2}{*}{personal health assistants} &
  access online medical information \cite{rodger2013pregnant} &
  information retrieval \\ \cline{3-4} 
 &
   &
  enable remote   healthcare \cite{zhou2018new} &
  speech recognition \\ \cline{2-4} 
 &
  \multirow{3}{3cm}{assisting the   elderly and the disabled} &
  daily assistance \cite{rani2017voice} &
  speech recognition, natural language understanding \\ \cline{3-4} 
 &
   &
  social interaction and company \cite{tapus2007socially,mavridis2015review} &
  speech recognition, speech synthesis \\ \cline{3-4} 
 &
   &
  assist people with speech   impairments \cite{green2021automatic,hux2017comprehension,hux2020effect,cassidy2016expressive,repova2020texttospeech,jreige2009vocalid},   hearing loss \cite{lyall2016smartphone}, dyslexia \cite{nittrouer2018speech},  or neurological disorders \cite{akbari2019reconstructing,anumanchipalli2019speech,herff2019generating} &
  speech recognition, speech synthesis \\ \hline
\multirow{6}{*}{Public Health} &
  \multirow{4}{3cm}{health knowledge popularization and medical education} &
  acquisition and representation of medical   knowledge \cite{goodwin2013automatic,rossanez2020kgen,rotmensch2017learning,riano2019ten,demner-fushman2007answering,jiang2021research,liu2017knowledgebased,terol2007knowledge,wang2017semantically} &
  knowledge engineering \\ \cline{3-4} 
 &
   &
  ease the access of medical   knowledge \cite{bao2020hhh,xu2018improve,mohan2018fast,peters2016translating,renato2018machine,tian2019smart,randhawa2013using} &
  question answering, information retrieval, machine translation \\ \cline{3-4} 
 &
   &
  generate medical case-based   questions \cite{leo2019ontologybased} &
  question generation \\ \cline{3-4} 
 &
   &
  construct simplified   summaries \cite{afzal2020clinical} &
  text summarization \\ \cline{2-4} 
 &
  \multirow{2}{*}{population screening} &
  identify target populations \cite{m2015state} &
  information extraction \\ \cline{3-4} 
 &
   &
  analyse of healthcare   questionnaire and surveys \cite{georgiou2015extracting} &
  information extraction \\ \hline
\multirow{7}{*}{Drug Development} &
  drug discovery &
  map the interactions between diseases, chemical compounds, and   biomacromolecules, predict molecular properties, and design novel   molecules \cite{ozturk2020exploring} &
  information extraction, information retrieval, knowledge   engineering \\ \cline{2-4} 
 &
  \multirow{2}{*}{preclinical research} &
  drug   screening \cite{lake2019artificial,pham2021deep,schwartz2013smifp} &
  information extraction \\ \cline{3-4} 
 &
   &
  predict adverse drug   reactions: side effect prediction \cite{zhang2021prediction}, and toxicity   prediction \cite{bouhedjar2020natural,jeon2019fp2vec} &
  information extraction \\ \cline{2-4} 
 &
  \multirow{3}{*}{clinical research} &
  clinical trial design \cite{nordon2019building} &
  information extraction, causal inference \\ \cline{3-4} 
 &
   &
  patient   recruitment \cite{chen2019clinical,harrer2019artificial,tissot2020natural} &
  information extraction \\ \cline{3-4} 
 &
   &
  clinical trial   analytics \cite{chen2020trends} &
  information extraction \\ \cline{2-4} 
 &
  drug review and safety   monitoring &
  adverse drug events discovery and drug safety   monitoring \cite{ventola2018big,wang2009active,liu2019drug} &
  information extraction \\ \hline
\end{tabular}%
}
\end{table*}

\section{NLP-driven Smart Healthcare for Specific Medical Issues}

NLP-driven smart healthcare plays an important role in many medical issues. In this section, we discuss how NLP-driven smart healthcare works in medical issues by taking two specific medical issues, i.e., COVID-19 pandemic and mental health, as examples.

\subsection{COVID-19 pandemic}

Worldwide outbreak of COVID-19 has triggered an unprecedented global health crisis and has attracted much attention from researchers \cite{zhou2022interpretable}. No wonder, the COVID-19 pandemic has become one of the most influential medical issues over the past few years. In the COVID-19 pandemic, NLP-driven smart healthcare can be utilized for pandemic prevention, diagnosing, and drug development.

Early forecasts of COVID-19 cases and pandemic knowledge popularization are crucial to the prevention of the COVID-19 pandemic. In \cite{zheng2020predicting}, an NLP module is embedded into an improved susceptible–infected model to build the proposed hybrid AI model for COVID-19 prediction, showing that the forecasting accuracy of COVID-19 cases can be improved by incorporating text inputs and with NLP techniques. In \cite{chen2021artificial}, the authors conclude that NLP techniques, e.g., NLP-aided information retrieval, literature-based discovery, question answering and etc., can be applied to address the information/knowledge needs of both researchers and the public in the COVID-19 pandemic.

In clinical practice, NLP can be utilized to identify positively diagnosed COVID19 patients from free text narratives \cite{chapman2020natural}, assess thoracic CT imaging reports \cite{cury2021natural}, and identify individuals with the greatest risk of severe complications due to COVID-19 \cite{decaprio2020building}, and provide COVID-19 testing advice \cite{meystre2021natural}. Such applications would be very useful to accelerate the diagnosis of COVID-19, mitigate its worst effects, and also reduce costs for combating the COVID-19 pandemic.

NLP has also been applied to drug development confronting COVID-19. In \cite{wang2022novel}, the authors developed an NLP method to automatically recognize the associations among potential targeted host organ systems, associated clinical manifestations and pathways, and suggest potential drug candidates. NLP models have also made great impacts in COVID-19 vaccine discovery through protein interaction prediction, molecular reaction modelling \cite{keshavarziarshadi2020artificial}. In addition, great opportunities for NLP can also be found in clinical design, regulatory decision-making, and pharmacovigilance \cite{liu2021aibased}. These applications would significantly reduce the time and cost of drug development for COVID-19.

\subsection{Mental health}
The mental health issues have received widespread and continuously increasing attention for many years. Specially, the World Health Organization (WHO) claimed that the pandemic and the resulting lockdowns, economic security, fear and uncertainty would further cause devastating impacts on people’s mental health the world over in the past several years \cite{who202010}. NLP-driven smart healthcare has great value in predicting/diagnosing and treating mental health conditions.

NLP techniques have been applied to early predict or identify/screen various mental disorders, such as psychiatric illness \cite{dai2021deepa}, late-life depression \cite{desouza2021natural}, severe mental illness (schizophrenia, schizoaffective disorder and bipolar disorder) \cite{jackson2017naturala}. In addition, some works have shown that NLP techniques can predict risk-taking behaviours (e.g., suicide) with good discrimination \cite{cohen2022integration,harvey2022natural} so that early interventions can be taken to save lives. The data collected for such analysis may include text data such as social media posts, screening surveys, EHRs \cite{zhang2022natural}, and also speech data come from narrative interviews \cite{bedi2015automated}, etc.

NLP techniques could also (automatically) provide effective psychotherapeutic interventions through web-based psycho-educational interventions, online counseling, etc., to augment therapist-based mental health interventions, showing potential future opportunities for their integration into online mental health tools \cite{calvo2017natural}. For example, the insights of \cite{althoff2016largescale} could help improve counselor training and generate real-time counseling quality monitoring and answer suggestion support tools. In addition, several mental health related areas that may benefit from NLP techniques, including characterizing and understanding mental disorders, measuring health outcomes, studying of social and occupational functioning, etc, were shown in \cite{harvey2022natural}. Specifically, \cite{chattaraman2019should} showed that the older would respond better to digital assistants employing a socially-oriented interaction style rather than the one with a task-oriented style, which is promising to promote mental health in older adults by providing social interaction and company.

\section{Limitations and Outlook}

Although recent advancements in deep learning and neural NLP have brought extraordinary enhancement to smart healthcare, there are still some limitations that current methods have yet to overcome.

\textbf{Understanding human language}. Although substantial efforts have been made to enable natural language understanding, the flexibility of human language still makes full understanding difficult, especially when ambiguity in biomedical texts is encountered. Misunderstanding could lead to inaccurate actions taken by robots, useless information returned by engines, and even wrong decisions made by decision support systems, leading to economic loss, time wasting, and even more serious consequences.

\textbf{Interpretability}. Although applications that rely on neural NLP to extract features and make decisions show excellent performance in real tasks, they are usually challenged by users due to their weakness in interpretability. Interpretability is essential for smart healthcare applications, especially in clinical scenarios that require quality assurance in cases of low confidence. One of the major interpretability issues is that the learned features are usually not understood by humans. In addition, when tuning pre-trained language models to downstream tasks, no enough intuitions on data for fine-tuning or types of applications can be given to guarantee good performance. Although efforts have been made to achieve interpretable NLP-driven applications, existing theories and methodologies are still not convincing and acceptable for many healthcare researchers and institutions. Before the interpretability issue is fully explored, the role of decision support systems in clinical practice can only be auxiliary from the perspectives of medical ethics and practical application.

\textbf{Implementation}. There are still many issues concerning the implementation of NLP-driven applications in smart healthcare. With the development of neural NLP, large deep neural networks (e.g., pre-trained language models) have been quickly migrated to smart healthcare. What followed are the increased requirements in computing power and training cost, and the concerns about the reliability of neural NLP systems. Patient privacy also prevents these models from achieving more prominent effects in smart healthcare for further practice. The consideration of medical ethics when applying such systems makes practical implementation more difficult.

In addition to tackling the aforementioned limitations, there are some other directions to enhance existing NLP systems for smart healthcare.

\textbf{Combining multiple NLP techniques}. One direction to enhance existing NLP systems can be the combination of multiple NLP techniques. For example, text generation can work as a data augmentation method for achieving comparable results in many applications with limited original data, such as training QA systems \cite{shen2019generation,wang2020texttosql} and other clinically relevant tasks \cite{amin-nejad2020exploring,ive2020generation}. Through automatic question generation, questionnaires and surveys for population screening can be generated from EHRs, which may outperform handcrafted ones. Machine translation has also proven beneficial for various text-based tasks by increasing the availability of multilingual healthcare information \cite{soto2019neural,wolk2015neuralbased,wolk2016translation}, implying the possibility of improving the performance of current CDS systems. In addition, exploration of general knowledge and domain knowledge in the field of NLP for smart healthcare deserves further attention and verification.

\textbf{End-to-end applications}. Current NLP driven applications for smart healthcare usually focus on dealing with tasks step by step and do not fully explore the feature extraction capability of advanced neural NLP for complex smart healthcare tasks. A deeper integration of NLP techniques and healthcare applications in an end-to-end manner can map the inputs and outputs directly, significantly simplify traditional pipelines for complex applications, eliminate the biases of intermediate components, and therefore achieve better performance. Taking population screening as an example, although NLP has been applied to identify populations and analyse screening test results in traditional screening procedures, NLP techniques can be further applied to build end-to-end population screening systems, with which the correlations between populations and optimal actions can be found to improve the screening performance and the quality of healthcare. Another example would be reducing the readmission rate. As mentioned before, some works have revealed that NLP has the ability to predict patient readmission, but further studies on providing appropriate interventions to reduce the readmission rate are not fully conducted. We look forward to studies that integrate the two parts to reveal every possibility for readmission rate reducing.

\textbf{Few-shot learning and incorporating domain knowledge}. By exploiting the learning capability of neural networks and large available corpora, neural NLP has shown powerful ability in learning language representations. However, for downstream tasks or smart healthcare applications, there is still a long way for NLP to go. Taking clinical decision support as an example, there are a lot of rare diseases with only a small number of observations available for training a clinical decision support system to distinguish rare diseases from common diseases. This is a quite challenging task, especially when there are similar outcomes among some rare diseases and common diseases. In addition, high-quality labelled data are undoubtedly essential to guarantee task accuracy in developing practical applications for smart healthcare. However, quality-controlled annotation not only requires a large amount of cost, but is also challenging due to the bias of experts' level of expertise. Therefore, even with well-learned pre-trained language models, few-shot learning algorithms and domain knowledge are expected to be applied so that the fine-tuned models would be effective in learning from few rare disease observations or limited high-quality labelled data.

\textbf{Incorporating multimodal and longitudinal data}. Finally, we also anticipate future intelligent systems to utilize all available AI techniques, not only NLP, for practical applications with high accuracy and reliability. The past few years have witnessed the dominance of data-driven approaches in many applications across various fields. NLP, computer vision, and other machine learning algorithms can be applied to analyse medical text, medical images, electronic recordings (e.g., heart sound), sensors data, laboratory results, and even genetic information. With multimodal learning, useful information extracted from these modalities can be combined together to perfectly fit the need for a complete and accurate analysis of available healthcare data and patients' health status. In addition, all of these data and clinical events can be longitudinal, where time series analysis can be applied to extract long-term dependencies and improve health care delivery. By combining these techniques to analyse multimodal and longitudinal data, future intelligent systems would become more powerful and reliable for patients, physicians, and healthcare institutions for applications such as 24/7 health monitoring, chronic-condition management, healthy lifestyle promotion, and precision medicine.

\section{Conclusion}
In the context of smart healthcare, NLP takes text or speech as the input in various scenarios involving humans and machines, and realizes the functions of analysing and understanding human language. In this paper, we review existing studies concerning NLP for smart healthcare from the perspectives of technique and application. We elaborate on different NLP approaches and the NLP pipeline for smart healthcare from the technical point of view. Table I provides the comparisons of different NLP approaches and their representative algorithms. Various text-oriented and speech-oriented NLP tasks are elaborated to conclude existing methodologies for tackling such tasks. By introducing smart healthcare applications employing NLP techniques in various smart healthcare scenarios (including clinical practice, hospital management, personal care, public health, and drug development), we show the strength and possibility of NLP for delivering smart healthcare. Table II provides a detailed list of representative applications in smart healthcare and their related NLP techniques. We further discuss two specific medical issues, i.e., COVID-19 pandemic and mental health, in which NLP-driven smart healthcare plays an important role. After that, we discuss the limitations of current works across understanding human language, interpretability, and implementation of NLP systems for smart healthcare. Finally, we identify several directions for future works, notably combining multiple NLP techniques, developing end-to-end applications, few-shot learning, and incorporating multimodal and longitudinal data.

\bibliographystyle{IEEEtran}
\bibliography{refer}

\begin{thebibliography}{100}
\providecommand{\url}[1]{#1}
\csname url@samestyle\endcsname
\providecommand{\newblock}{\relax}
\providecommand{\bibinfo}[2]{#2}
\providecommand{\BIBentrySTDinterwordspacing}{\spaceskip=0pt\relax}
\providecommand{\BIBentryALTinterwordstretchfactor}{4}
\providecommand{\BIBentryALTinterwordspacing}{\spaceskip=\fontdimen2\font plus
\BIBentryALTinterwordstretchfactor\fontdimen3\font minus
  \fontdimen4\font\relax}
\providecommand{\BIBforeignlanguage}[2]{{%
\expandafter\ifx\csname l@#1\endcsname\relax
\typeout{** WARNING: IEEEtran.bst: No hyphenation pattern has been}%
\typeout{** loaded for the language `#1'. Using the pattern for}%
\typeout{** the default language instead.}%
\else
\language=\csname l@#1\endcsname
\fi
#2}}
\providecommand{\BIBdecl}{\relax}
\BIBdecl

\bibitem{tian2019smart}
S.~Tian, W.~Yang, J.~M.~L. Grange, P.~Wang, W.~Huang, and Z.~Ye,
  ``\BIBforeignlanguage{en}{Smart healthcare: Making medical care more
  intelligent},'' \emph{\BIBforeignlanguage{en}{Global Health Journal}},
  vol.~3, no.~3, pp. 62--65, Sep. 2019.

\bibitem{young2018recent}
T.~Young, D.~Hazarika, S.~Poria, and E.~Cambria, ``Recent trends in deep
  learning based natural language processing,'' \emph{arXiv:1708.02709 [cs]},
  Nov. 2018.

\bibitem{bengio2003neural}
Y.~Bengio, R.~Ducharme, P.~Vincent, and C.~Jauvin, ``A neural probabilistic
  language model,'' \emph{Journal of Machine Learning Research}, vol.~3, no.
  Feb, pp. 1137--1155, 2003.

\bibitem{crim2005automatically}
J.~Crim, R.~McDonald, and F.~Pereira, ``Automatically annotating documents with
  normalized gene lists,'' \emph{BMC Bioinformatics}, vol.~6, no.~1, p. S13,
  May 2005.

\bibitem{vilares2008extraction}
J.~Vilares, M.~A. Alonso, and M.~Vilares, ``Extraction of complex index terms
  in non-{{English IR}}: {{A}} shallow parsing based approach,''
  \emph{Information Processing \& Management}, vol.~44, no.~4, pp. 1517--1537,
  Jul. 2008.

\bibitem{chiticariu2013rulebased}
L.~Chiticariu, Y.~Li, and F.~R. Reiss, ``Rule-based information extraction is
  dead! {{Long}} live rule-based information extraction systems!'' in
  \emph{Proceedings of the 2013 {{Conference}} on {{Empirical Methods}} in
  {{Natural Language Processing}}}.\hskip 1em plus 0.5em minus 0.4em\relax
  {Seattle, Washington, USA}: {Association for Computational Linguistics}, Oct.
  2013, pp. 827--832.

\bibitem{kang2012using}
N.~Kang, B.~Singh, Z.~Afzal, E.~M. {van Mulligen}, and J.~Kors, ``Using
  rule-based natural language processing to improve disease normalization in
  biomedical text,'' \emph{Journal of the American Medical Informatics
  Association : JAMIA}, vol.~20, Oct. 2012.

\bibitem{weng2017medical}
W.-H. Weng, K.~B. Wagholikar, A.~T. McCray, P.~Szolovits, and H.~C. Chueh,
  ``Medical subdomain classification of clinical notes using a machine
  learning-based natural language processing approach,'' \emph{BMC Medical
  Informatics and Decision Making}, vol.~17, p. 155, Dec. 2017.

\bibitem{dessi2020tfidf}
D.~Dessi, R.~Helaoui, V.~Kumar, D.~R. Recupero, and D.~Riboni, ``{{TF}}-{{IDF}}
  vs word embeddings for morbidity identification in clinical notes: An initial
  study,'' \emph{arXiv:2105.09632 [cs]}, Mar. 2020.

\bibitem{ozyegen2021wordlevel}
O.~Ozyegen, D.~Kabe, and M.~Cevik, ``Word-level text highlighting of medical
  texts {{forTelehealth}} services,'' \emph{arXiv:2105.10400 [cs]}, May 2021.

\bibitem{rahimian2019significant}
M.~Rahimian, J.~L. Warner, S.~K. Jain, R.~B. Davis, J.~A. Zerillo, and R.~M.
  Joyce, ``Significant and distinctive n-grams in oncology notes: {{A}}
  text-mining method to analyze the effect of {{OpenNotes}} on clinical
  documentation,'' \emph{JCO Clinical Cancer Informatics}, no.~3, pp. 1--9,
  Dec. 2019.

\bibitem{yazdani2019words}
A.~Yazdani, R.~Safdari, A.~Golkar, and S.~Rostam Niakan~Kalhori, ``Words
  prediction based on {{N}}-gram model for free-text entry in electronic health
  records,'' \emph{Health Information Science and Systems}, vol.~7, Feb. 2019.

\bibitem{yip2010concept}
V.~Yip, M.~Mete, U.~Topaloglu, and S.~Kockara, ``Concept discovery for
  pathology reports using an {{N}}-gram model,'' \emph{Summit on Translational
  Bioinformatics}, vol. 2010, pp. 43--47, Mar. 2010.

\bibitem{beeksma2019predicting}
M.~Beeksma, S.~Verberne, A.~{van den Bosch}, E.~Das, I.~Hendrickx, and
  S.~Groenewoud, ``Predicting life expectancy with a long short-term memory
  recurrent neural network using electronic medical records,'' \emph{BMC
  Medical Informatics and Decision Making}, vol.~19, no.~1, p.~36, Feb. 2019.

\bibitem{jagannatha2016bidirectional}
A.~N. Jagannatha and H.~Yu, ``Bidirectional {{RNN}} for medical event detection
  in electronic health records,'' in \emph{Proceedings of the 2016
  {{Conference}} of the {{North American Chapter}} of the {{Association}} for
  {{Computational Linguistics}}: {{Human Language Technologies}}}.\hskip 1em
  plus 0.5em minus 0.4em\relax {San Diego, California}: {Association for
  Computational Linguistics}, Jun. 2016, pp. 473--482.

\bibitem{liu2016augmented}
C.~Liu, H.~Sun, N.~Du, S.~Tan, H.~Fei, W.~Fan, T.~Yang, H.~Wu, Y.~Li, and
  C.~Zhang, ``Augmented {{LSTM}} framework to construct medical self-diagnosis
  android,'' in \emph{2016 {{IEEE}} 16th {{International Conference}} on {{Data
  Mining}} ({{ICDM}})}, Dec. 2016, pp. 251--260.

\bibitem{zhao2018leveraging}
Y.-S. Zhao, K.-L. Zhang, H.-C. Ma, and K.~Li, ``Leveraging text skeleton for
  de-identification of electronic medical records,'' \emph{BMC Medical
  Informatics and Decision Making}, vol.~18, no. Suppl 1, p.~18, Mar. 2018.

\bibitem{hughes2017medical}
M.~Hughes, I.~Li, S.~Kotoulas, and T.~Suzumura,
  ``\BIBforeignlanguage{eng}{Medical text classification using convolutional
  neural networks},'' \emph{\BIBforeignlanguage{eng}{Studies in Health
  Technology and Informatics}}, vol. 235, pp. 246--250, 2017.

\bibitem{li2019intelligent}
X.~Li, H.~Wang, H.~He, J.~Du, J.~Chen, and J.~Wu, ``Intelligent diagnosis with
  chinese electronic medical records based on convolutional neural networks,''
  \emph{BMC Bioinformatics}, vol.~20, no.~1, p.~62, Feb. 2019.

\bibitem{li2020graph}
Y.~Li, B.~Qian, X.~Zhang, and H.~Liu, ``Graph neural network-based diagnosis
  prediction,'' \emph{Big Data}, vol.~8, no.~5, pp. 379--390, Oct. 2020.

\bibitem{sun2021disease}
Z.~Sun, H.~Yin, H.~Chen, T.~Chen, L.~Cui, and F.~Yang, ``Disease prediction via
  graph neural networks,'' \emph{IEEE Journal of Biomedical and Health
  Informatics}, vol.~25, no.~3, pp. 818--826, Mar. 2021.

\bibitem{wu2021leveraging}
T.~Wu, Y.~Wang, Y.~Wang, E.~Zhao, and Y.~Yuan,
  ``\BIBforeignlanguage{en}{Leveraging graph-based hierarchical medical entity
  embedding for healthcare applications},''
  \emph{\BIBforeignlanguage{en}{Scientific Reports}}, vol.~11, no.~1, p. 5858,
  Mar. 2021.

\bibitem{mayer2020transformerbased}
T.~Mayer, E.~Cabrio, and S.~Villata, ``Transformer-based argument mining for
  healthcare applications,'' in \emph{{{ECAI}} 2020 - 24th {{European
  Conference}} on {{Artificial Intelligence}}}, {Santiago de Compostela /
  Online, Spain}, Aug. 2020.

\bibitem{zhang2020timeaware}
D.~Zhang, J.~Thadajarassiri, C.~Sen, and E.~Rundensteiner,
  ``\BIBforeignlanguage{en}{Time-aware transformer-based network for clinical
  notes series prediction},'' in \emph{\BIBforeignlanguage{en}{Machine
  {{Learning}} for {{Healthcare Conference}}}}.\hskip 1em plus 0.5em minus
  0.4em\relax {PMLR}, Sep. 2020, pp. 566--588.

\bibitem{tokala2018deep}
S.~Tokala, V.~Gambhir, and A.~Mukherjee, ``Deep learning for social media
  health text classification,'' in \emph{Proceedings of the 2018 {{EMNLP
  Workshop SMM4H}}: {{The}} 3rd {{Social Media Mining}} for {{Health
  Applications Workshop}} \& {{Shared Task}}}.\hskip 1em plus 0.5em minus
  0.4em\relax {Brussels, Belgium}: {Association for Computational Linguistics},
  Oct. 2018, pp. 61--64.

\bibitem{choi2017retain}
E.~Choi, M.~T. Bahadori, J.~A. Kulas, A.~Schuetz, W.~F. Stewart, and J.~Sun,
  ``{{RETAIN}}: An interpretable predictive model for healthcare using reverse
  time attention mechanism,'' \emph{arXiv:1608.05745 [cs]}, Feb. 2017.

\bibitem{chu2018using}
J.~Chu, W.~Dong, K.~He, H.~Duan, and Z.~Huang, ``\BIBforeignlanguage{en}{Using
  neural attention networks to detect adverse medical events from electronic
  health records},'' \emph{\BIBforeignlanguage{en}{Journal of Biomedical
  Informatics}}, vol.~87, pp. 118--130, Nov. 2018.

\bibitem{chakraborty2020explicitblurred}
P.~Chakraborty, F.~Wang, J.~Hu, and D.~Sow, ``Explicit-blurred memory network
  for analyzing patient electronic health records,'' \emph{arXiv:1911.06472
  [cs, stat]}, Jul. 2020.

\bibitem{song2021local}
J.~Song, Y.~Wang, S.~Tang, Y.~Zhang, Z.~Chen, Z.~Zhang, T.~Zhang, and F.~Wu,
  ``Local\textendash{{Global}} memory neural network for medication
  prediction,'' \emph{IEEE Transactions on Neural Networks and Learning
  Systems}, vol.~32, no.~4, pp. 1723--1736, Apr. 2021.

\bibitem{yoon2019information}
H.-J. Yoon, J.~Gounley, M.~T. Young, and G.~Tourassi, ``Information extraction
  from cancer pathology reports with graph convolution networks for natural
  language texts,'' in \emph{2019 {{IEEE International Conference}} on {{Big
  Data}} ({{Big Data}})}, Dec. 2019, pp. 4561--4564.

\bibitem{cai2017cnnlstm}
R.~Cai, B.~Zhu, L.~Ji, T.~Hao, J.~Yan, and W.~Liu, ``An {{CNN}}-{{LSTM}}
  attention approach to understanding user query intent from online health
  communities,'' in \emph{2017 {{IEEE International Conference}} on {{Data
  Mining Workshops}} ({{ICDMW}})}, Nov. 2017, pp. 430--437.

\bibitem{tang2019entity}
B.~Tang, X.~Wang, J.~Yan, and Q.~Chen, ``Entity recognition in chinese clinical
  text using attention-based {{CNN}}-{{LSTM}}-{{CRF}},'' \emph{BMC Medical
  Informatics and Decision Making}, vol.~19, no.~3, p.~74, Apr. 2019.

\bibitem{choi2020learning}
E.~Choi, Z.~Xu, Y.~Li, M.~W. Dusenberry, G.~Flores, Y.~Xue, and A.~M. Dai,
  ``Learning the graphical structure of electronic health records with graph
  convolutional transformer,'' \emph{arXiv:1906.04716 [cs, stat]}, Jan. 2020.

\bibitem{wang2020documentlevel}
J.~Wang, X.~Chen, Y.~Zhang, Y.~Zhang, J.~Wen, H.~Lin, Z.~Yang, and X.~Wang,
  ``Document-level biomedical relation extraction using graph convolutional
  network and multihead attention: Algorithm development and validation,''
  \emph{JMIR Medical Informatics}, vol.~8, no.~7, p. e17638, Jul. 2020.

\bibitem{qiu2020pretrained}
X.~Qiu, T.~Sun, Y.~Xu, Y.~Shao, N.~Dai, and X.~Huang, ``Pre-trained models for
  natural language processing: {{A}} survey,'' \emph{arXiv:2003.08271 [cs]},
  Apr. 2020.

\bibitem{beam2019clinical}
A.~L. Beam, B.~Kompa, A.~Schmaltz, I.~Fried, G.~Weber, N.~P. Palmer, X.~Shi,
  T.~Cai, and I.~S. Kohane, ``Clinical concept embeddings learned from massive
  sources of multimodal medical data,'' \emph{arXiv:1804.01486 [cs, stat]},
  Aug. 2019.

\bibitem{cai2018medical}
X.~Cai, J.~Gao, K.~Y. Ngiam, B.~C. Ooi, Y.~Zhang, and X.~Yuan, ``Medical
  concept embedding with time-aware attention,'' in \emph{Proceedings of the
  27th {{International Joint Conference}} on {{Artificial Intelligence}}}, ser.
  {{IJCAI}}'18.\hskip 1em plus 0.5em minus 0.4em\relax {Stockholm, Sweden}:
  {AAAI Press}, Jul. 2018, pp. 3984--3990.

\bibitem{kholghi2016benefits}
M.~Kholghi, L.~De~Vine, L.~Sitbon, G.~Zuccon, and A.~Nguyen, ``The benefits of
  word embeddings features for active learning in clinical information
  extraction,'' in \emph{Proceedings of the {{Australasian Language Technology
  Association Workshop}} 2016}, {Melbourne, Australia}, Dec. 2016, pp. 25--34.

\bibitem{zhang2019biowordvec}
Y.~Zhang, Q.~Chen, Z.~Yang, H.~Lin, and Z.~Lu,
  ``\BIBforeignlanguage{en}{{{BioWordVec}}, improving biomedical word
  embeddings with subword information and {{MeSH}}},''
  \emph{\BIBforeignlanguage{en}{Scientific Data}}, vol.~6, no.~1, p.~52, May
  2019.

\bibitem{dubois2018effective}
S.~Dubois, N.~Romano, D.~C. Kale, N.~Shah, and K.~Jung, ``Effective
  representations of clinical notes,'' \emph{arXiv:1705.07025 [cs, stat]}, Aug.
  2018.

\bibitem{jin2019probing}
Q.~Jin, B.~Dhingra, W.~W. Cohen, and X.~Lu, ``Probing biomedical embeddings
  from language models,'' \emph{arXiv:1904.02181 [cs]}, Apr. 2019.

\bibitem{lee2020biobert}
J.~Lee, W.~Yoon, S.~Kim, D.~Kim, S.~Kim, C.~H. So, and J.~Kang, ``{{BioBERT}}:
  A pre-trained biomedical language representation model for biomedical text
  mining,'' \emph{Bioinformatics}, vol.~36, no.~4, pp. 1234--1240, Feb. 2020.

\bibitem{huang2020clinicalbert}
K.~Huang, J.~Altosaar, and R.~Ranganath, ``{{ClinicalBERT}}: Modeling clinical
  notes and predicting hospital readmission,'' \emph{arXiv:1904.05342 [cs]},
  Nov. 2020.

\bibitem{rasmy2021medbert}
L.~Rasmy, Y.~Xiang, Z.~Xie, C.~Tao, and D.~Zhi,
  ``\BIBforeignlanguage{en}{Med-{{BERT}}: Pretrained contextualized embeddings
  on large-scale structured electronic health records for disease
  prediction},'' \emph{\BIBforeignlanguage{en}{npj Digital Medicine}}, vol.~4,
  no.~1, pp. 1--13, May 2021.

\bibitem{li2020behrt}
Y.~Li, S.~Rao, J.~R.~A. Solares, A.~Hassaine, R.~Ramakrishnan, D.~Canoy,
  Y.~Zhu, K.~Rahimi, and G.~{Salimi-Khorshidi},
  ``\BIBforeignlanguage{en}{{{BEHRT}}: Transformer for electronic health
  records},'' \emph{\BIBforeignlanguage{en}{Scientific Reports}}, vol.~10,
  no.~1, p. 7155, Apr. 2020.

\bibitem{schneider2021gpt2}
E.~T.~R. Schneider, J.~V.~A. {de Souza}, Y.~B. Gumiel, C.~Moro, and E.~C.
  Paraiso, ``A {{GPT}}-2 language model for biomedical texts in portuguese,''
  in \emph{2021 {{IEEE}} 34th {{International Symposium}} on
  {{Computer}}-{{Based Medical Systems}} ({{CBMS}})}, Jun. 2021, pp. 474--479.

\bibitem{akkasi2016chemtok}
A.~Akkasi, E.~Varo{\u g}lu, and N.~Dimililer, ``{{ChemTok}}: {{A}} new rule
  based tokenizer for chemical named entity recognition,'' \emph{BioMed
  Research International}, vol. 2016, 2016.

\bibitem{dai2015enhancing}
H.-J. Dai, P.-T. Lai, Y.-C. Chang, and R.~T.-H. Tsai,
  ``\BIBforeignlanguage{eng}{Enhancing of chemical compound and drug name
  recognition using representative tag scheme and fine-grained tokenization},''
  \emph{\BIBforeignlanguage{eng}{Journal of Cheminformatics}}, vol.~7, no.
  Suppl 1 Text mining for chemistry and the CHEMDNER track, p. S14, 2015.

\bibitem{liu2012biolemmatizer}
H.~Liu, T.~Christiansen, W.~A. Baumgartner, and K.~Verspoor,
  ``{{BioLemmatizer}}: A lemmatization tool for morphological processing of
  biomedical text,'' \emph{Journal of Biomedical Semantics}, vol.~3, p.~3, Apr.
  2012.

\bibitem{qiao2019understanding}
Y.~Qiao, C.~Xiong, Z.~Liu, and Z.~Liu, ``Understanding the {{Behaviors}} of
  {{BERT}} in {{Ranking}},'' \emph{arXiv:1904.07531 [cs]}, Apr. 2019.

\bibitem{liu2021selfsupervised}
X.~Liu, F.~Zhang, Z.~Hou, Z.~Wang, L.~Mian, J.~Zhang, and J.~Tang,
  ``Self-supervised learning: Generative or contrastive,'' \emph{IEEE
  Transactions on Knowledge and Data Engineering}, pp. 1--1, 2021.

\bibitem{wang2019clinical}
Y.~Wang, S.~Sohn, S.~Liu, F.~Shen, L.~Wang, E.~J. Atkinson, S.~Amin, and
  H.~Liu, ``A clinical text classification paradigm using weak supervision and
  deep representation,'' \emph{BMC Medical Informatics and Decision Making},
  vol.~19, no.~1, p.~1, Jan. 2019.

\bibitem{hassanpour2016information}
S.~Hassanpour and C.~P. Langlotz, ``Information extraction from
  multi-institutional radiology reports,'' \emph{Artificial intelligence in
  medicine}, vol.~66, pp. 29--39, Jan. 2016.

\bibitem{perera2020named}
N.~Perera, M.~Dehmer, and F.~{Emmert-Streib}, ``Named entity recognition and
  relation detection for biomedical information extraction,'' \emph{Frontiers
  in Cell and Developmental Biology}, vol.~8, p. 673, Aug. 2020.

\bibitem{thillaisundaram2019biomedical}
A.~Thillaisundaram and T.~Togia, ``Biomedical relation extraction with
  pre-trained language representations and minimal task-specific
  architecture,'' in \emph{Proceedings of {{The}} 5th {{Workshop}} on {{BioNLP
  Open Shared Tasks}}}.\hskip 1em plus 0.5em minus 0.4em\relax {Hong Kong,
  China}: {Association for Computational Linguistics}, Nov. 2019, pp. 84--89.

\bibitem{zitnik2015sievebased}
S.~{\v Z}itnik, M.~{\v Z}itnik, B.~Zupan, and M.~Bajec, ``Sieve-based relation
  extraction of gene regulatory networks from biological literature,''
  \emph{BMC Bioinformatics}, vol.~16, no. Suppl 16, p.~S1, Oct. 2015.

\bibitem{jindal2013extraction}
P.~Jindal and D.~Roth, ``\BIBforeignlanguage{en}{Extraction of events and
  temporal expressions from clinical narratives},''
  \emph{\BIBforeignlanguage{en}{Journal of Biomedical Informatics}}, vol.~46,
  pp. S13--S19, Dec. 2013.

\bibitem{garg2018machine}
A.~Garg and M.~Agarwal, ``Machine translation: A literature review,''
  \emph{arXiv:1901.01122 [cs]}, Dec. 2018.

\bibitem{berard2020multilingual}
A.~B{\'e}rard, Z.~M. Kim, V.~Nikoulina, E.~L. Park, and M.~Gall{\'e}, ``A
  multilingual neural machine translation model for biomedical data,'' in
  \emph{Proceedings of the 1st {{Workshop}} on {{NLP}} for {{COVID}}-19
  ({{Part}} 2) at {{EMNLP}} 2020}.\hskip 1em plus 0.5em minus 0.4em\relax
  {Online}: {Association for Computational Linguistics}, Dec. 2020.

\bibitem{kirchhoff2011application}
K.~Kirchhoff, A.~M. Turner, A.~Axelrod, and F.~Saavedra, ``Application of
  statistical machine translation to public health information: A feasibility
  study,'' \emph{Journal of the American Medical Informatics Association :
  JAMIA}, vol.~18, no.~4, pp. 473--478, 2011.

\bibitem{afzal2020clinical}
M.~Afzal, F.~Alam, K.~M. Malik, and G.~M. Malik,
  ``\BIBforeignlanguage{EN}{Clinical {{Context}}\textendash{{Aware}} biomedical
  text summarization using deep neural network: Model development and
  validation},'' \emph{\BIBforeignlanguage{EN}{Journal of Medical Internet
  Research}}, vol.~22, no.~10, p. e19810, Oct. 2020.

\bibitem{lopez2019automatic}
J.~Lopez, ``Automatic summarization of medical conversations, a review,'' in
  \emph{{{TALN}}-{{RECITAL}} 2019-{{PFIA}} 2019}.\hskip 1em plus 0.5em minus
  0.4em\relax {Toulouse, France}: {ATALA}, Jul. 2019, pp. 487--498.

\bibitem{manas2021knowledgeinfused}
G.~Manas, V.~Aribandi, U.~Kursuncu, A.~Alambo, V.~L. Shalin, K.~Thirunarayan,
  J.~Beich, M.~Narasimhan, and A.~Sheth,
  ``\BIBforeignlanguage{EN}{Knowledge-infused abstractive summarization of
  clinical diagnostic interviews: Framework development study},''
  \emph{\BIBforeignlanguage{EN}{JMIR Mental Health}}, vol.~8, no.~5, p. e20865,
  May 2021.

\bibitem{pistol2018meditest}
I.~Pistol, D.~Trandab{\u a}ț, and M.~R{\u a}schip,
  ``\BIBforeignlanguage{en}{Medi-test: Generating tests from medical reference
  texts},'' \emph{\BIBforeignlanguage{en}{Data}}, vol.~3, no.~4, p.~70, Dec.
  2018.

\bibitem{shen2019generation}
S.~Shen, Y.~Li, N.~Du, X.~Wu, Y.~Xie, S.~Ge, T.~Yang, K.~Wang, X.~Liang, and
  W.~Fan, ``On the generation of medical question-answer pairs,''
  \emph{arXiv:1811.00681 [cs]}, Dec. 2019.

\bibitem{wang2008automatic}
W.~Wang, T.~Hao, and W.~Liu, ``\BIBforeignlanguage{en}{Automatic question
  generation for learning evaluation in medicine},'' in
  \emph{\BIBforeignlanguage{en}{Advances in {{Web Based Learning}}
  \textendash{} {{ICWL}} 2007}}, ser. Lecture {{Notes}} in {{Computer
  Science}}, H.~Leung, F.~Li, R.~Lau, and Q.~Li, Eds.\hskip 1em plus 0.5em
  minus 0.4em\relax {Berlin, Heidelberg}: {Springer}, 2008, pp. 242--251.

\bibitem{li2021semisupervised}
D.~Li, Z.~Ren, P.~Ren, Z.~Chen, M.~Fan, J.~Ma, and M.~{de Rijke},
  ``Semi-supervised variational reasoning for medical dialogue generation,''
  \emph{Proceedings of the 44th International ACM SIGIR Conference on Research
  and Development in Information Retrieval}, pp. 544--554, Jul. 2021.

\bibitem{lin2020graphevolving}
S.~Lin, P.~Zhou, X.~Liang, J.~Tang, R.~Zhao, Z.~Chen, and L.~Lin,
  ``Graph-evolving meta-learning for low-resource medical dialogue
  generation,'' \emph{arXiv:2012.11988 [cs]}, Dec. 2020.

\bibitem{yang2020generation}
W.~Yang, G.~Zeng, B.~Tan, Z.~Ju, S.~Chakravorty, X.~He, S.~Chen, X.~Yang,
  Q.~Wu, Z.~Yu, E.~Xing, and P.~Xie, ``On the generation of medical dialogues
  for {{COVID}}-19,'' \emph{arXiv:2005.05442 [cs]}, Jun. 2020.

\bibitem{pauws2018making}
S.~Pauws, A.~Gatt, E.~Krahmer, and E.~Reiter, ``Making effective use of
  healthcare data using data-to-text technology,'' \emph{arXiv:1808.03507
  [cs]}, Aug. 2018.

\bibitem{kougia2019survey}
V.~Kougia, J.~Pavlopoulos, and I.~Androutsopoulos, ``A survey on biomedical
  image captioning,'' \emph{arXiv:1905.13302 [cs]}, May 2019.

\bibitem{xiong2019reinforced}
Y.~Xiong, B.~Du, and P.~Yan, ``\BIBforeignlanguage{en}{Reinforced transformer
  for medical image captioning},'' in \emph{\BIBforeignlanguage{en}{Machine
  {{Learning}} in {{Medical Imaging}}}}, ser. Lecture {{Notes}} in {{Computer
  Science}}, H.-I. Suk, M.~Liu, P.~Yan, and C.~Lian, Eds.\hskip 1em plus 0.5em
  minus 0.4em\relax {Cham}: {Springer International Publishing}, 2019, pp.
  673--680.

\bibitem{he2021visual}
X.~He, Z.~Cai, W.~Wei, Y.~Zhang, L.~Mou, E.~Xing, and P.~Xie, ``Towards visual
  question answering on pathology images,'' in \emph{Proceedings of the 59th
  {{Annual Meeting}} of the {{Association}} for {{Computational Linguistics}}
  and the 11th {{International Joint Conference}} on {{Natural Language
  Processing}} ({{Volume}} 2: {{Short Papers}})}.\hskip 1em plus 0.5em minus
  0.4em\relax {Online}: {Association for Computational Linguistics}, Aug. 2021,
  pp. 708--718.

\bibitem{zhan2020medical}
L.-M. Zhan, B.~Liu, L.~Fan, J.~Chen, and X.-M. Wu, ``Medical visual question
  answering via conditional reasoning,'' in \emph{Proceedings of the 28th {{ACM
  International Conference}} on {{Multimedia}}}, ser. {{MM}} '20.\hskip 1em
  plus 0.5em minus 0.4em\relax {New York, NY, USA}: {Association for Computing
  Machinery}, Oct. 2020, pp. 2345--2354.

\bibitem{afrae2019new}
B.~Afrae, D.~Yousra, A.~Imane, B.~A. Mohamed, and A.~B. Abdelhakim, ``A new
  visual question answering system for medical images characterization,'' in
  \emph{Proceedings of the 4th {{International Conference}} on {{Smart City
  Applications}}}, ser. {{SCA}} '19.\hskip 1em plus 0.5em minus 0.4em\relax
  {New York, NY, USA}: {Association for Computing Machinery}, Oct. 2019, pp.
  1--7.

\bibitem{wu2018semehr}
H.~Wu, G.~Toti, K.~I. Morley, Z.~M. Ibrahim, A.~Folarin, R.~Jackson,
  I.~Kartoglu, A.~Agrawal, C.~Stringer, D.~Gale, G.~Gorrell, A.~Roberts,
  M.~Broadbent, R.~Stewart, and R.~J. Dobson, ``{{SemEHR}}: {{A}}
  general-purpose semantic search system to surface semantic data from clinical
  notes for tailored care, trial recruitment, and clinical research,''
  \emph{Journal of the American Medical Informatics Association : JAMIA},
  vol.~25, no.~5, pp. 530--537, Jan. 2018.

\bibitem{gobeill2015deep}
J.~Gobeill, A.~Gaudinat, E.~Pasche, D.~Vishnyakova, P.~Gaudet, A.~Bairoch, and
  P.~Ruch, ``Deep question answering for protein annotation,'' \emph{Database:
  The Journal of Biological Databases and Curation}, vol. 2015, Sep. 2015.

\bibitem{montazeralghaem2020relevance}
A.~Montazeralghaem, R.~Rahimi, and J.~Allan, ``Relevance ranking based on
  query-aware context analysis,'' \emph{Advances in Information Retrieval},
  vol. 12035, pp. 446--460, Mar. 2020.

\bibitem{xu2018improve}
B.~Xu, H.~Lin, Y.~Lin, Y.~Ma, L.~Yang, J.~Wang, and Z.~Yang, ``Improve
  biomedical information retrieval using modified learning to rank methods,''
  \emph{IEEE/ACM Transactions on Computational Biology and Bioinformatics},
  vol.~15, no.~6, pp. 1797--1809, Nov. 2018.

\bibitem{urbain2009passage}
J.~Urbain, O.~Frieder, and N.~Goharian, ``Passage relevance models for genomics
  search,'' \emph{BMC Bioinformatics}, vol.~10, no. Suppl 3, p.~S3, Mar. 2009.

\bibitem{mohan2018fast}
S.~Mohan, N.~Fiorini, S.~Kim, and Z.~Lu, ``A fast deep learning model for
  textual relevance in biomedical information retrieval,'' in \emph{Proceedings
  of the 2018 {{World Wide Web Conference}}}, ser. {{WWW}} '18.\hskip 1em plus
  0.5em minus 0.4em\relax {Republic and Canton of Geneva, CHE}: {International
  World Wide Web Conferences Steering Committee}, Apr. 2018, pp. 77--86.

\bibitem{hristovski2015biomedical}
D.~Hristovski, D.~Dinevski, A.~Kastrin, and T.~C. Rindflesch, ``Biomedical
  question answering using semantic relations,'' \emph{BMC Bioinformatics},
  vol.~16, no.~1, p.~6, Jan. 2015.

\bibitem{vinod2021multilingual}
V.~Vinod, S.~Agrawal, V.~Gaurav, P.~R, and S.~Choudhary, ``Multilingual medical
  question answering and information retrieval for rural health intelligence
  access,'' \emph{arXiv:2106.01251 [cs]}, Jun. 2021.

\bibitem{zahid2018cliniqa}
M.~A.~H. Zahid, A.~Mittal, R.~C. Joshi, and G.~Atluri, ``{{CLINIQA}}: A machine
  intelligence based clinical question answering system,''
  \emph{arXiv:1805.05927 [cs]}, May 2018.

\bibitem{wang2020texttosql}
P.~Wang, T.~Shi, and C.~K. Reddy, ``\BIBforeignlanguage{en}{Text-to-{{SQL}}
  generation for question answering on electronic medical records},'' in
  \emph{\BIBforeignlanguage{en}{Proceedings of {{The Web Conference}}
  2020}}.\hskip 1em plus 0.5em minus 0.4em\relax {Taipei Taiwan}: {ACM}, Apr.
  2020, pp. 350--361.

\bibitem{jiang2021research}
Z.~Jiang, C.~Chi, and Y.~Zhan, ``Research on medical question answering system
  based on knowledge graph,'' \emph{IEEE Access}, vol.~9, pp. 21\,094--21\,101,
  2021.

\bibitem{liu2017knowledgebased}
H.~Liu, Q.~Hu, Y.~Zhang, C.~Xing, and M.~Sheng, ``\BIBforeignlanguage{en}{A
  knowledge-based health question answering system},'' in
  \emph{\BIBforeignlanguage{en}{Smart {{Health}}}}, ser. Lecture {{Notes}} in
  {{Computer Science}}, H.~Chen, D.~D. Zeng, E.~Karahanna, and I.~Bardhan,
  Eds.\hskip 1em plus 0.5em minus 0.4em\relax {Cham}: {Springer International
  Publishing}, 2017, pp. 286--291.

\bibitem{demner-fushman2007answering}
D.~{Demner-Fushman} and J.~Lin, ``\BIBforeignlanguage{en}{Answering clinical
  questions with knowledge-based and statistical techniques},''
  \emph{\BIBforeignlanguage{en}{Computational Linguistics}}, vol.~33, no.~1,
  pp. 63--103, Mar. 2007.

\bibitem{terol2007knowledge}
R.~M. Terol, P.~{Mart{\'i}nez-Barco}, and M.~Palomar,
  ``\BIBforeignlanguage{en}{A knowledge based method for the medical question
  answering problem},'' \emph{\BIBforeignlanguage{en}{Computers in Biology and
  Medicine}}, vol.~37, no.~10, pp. 1511--1521, Oct. 2007.

\bibitem{liu2018tknow}
Z.~Liu, E.~Peng, S.~Yan, G.~Li, and T.~Hao, ``T-know: A knowledge graph-based
  question answering and infor-mation retrieval system for traditional chinese
  medicine,'' in \emph{Proceedings of the 27th {{International Conference}} on
  {{Computational Linguistics}}: {{System Demonstrations}}}.\hskip 1em plus
  0.5em minus 0.4em\relax {Santa Fe, New Mexico}: {Association for
  Computational Linguistics}, Aug. 2018, pp. 15--19.

\bibitem{mutabazi2021review}
E.~Mutabazi, J.~Ni, G.~Tang, and W.~Cao, ``\BIBforeignlanguage{en}{A review on
  medical textual question answering systems based on deep learning
  approaches},'' \emph{\BIBforeignlanguage{en}{Applied Sciences}}, vol.~11,
  no.~12, p. 5456, Jan. 2021.

\bibitem{shim2021building}
H.~Shim, D.~Lowet, S.~Luca, and B.~Vanrumste, ``Building blocks of a
  task-oriented dialogue system in the healthcare domain,'' in
  \emph{Proceedings of the {{Second Workshop}} on {{Natural Language
  Processing}} for {{Medical Conversations}}}.\hskip 1em plus 0.5em minus
  0.4em\relax {Online}: {Association for Computational Linguistics}, Jun. 2021,
  pp. 47--57.

\bibitem{wei2018taskoriented}
Z.~Wei, Q.~Liu, B.~Peng, H.~Tou, T.~Chen, X.~Huang, K.-f. Wong, and X.~Dai,
  ``Task-oriented dialogue system for automatic diagnosis,'' in
  \emph{Proceedings of the 56th {{Annual Meeting}} of the {{Association}} for
  {{Computational Linguistics}} ({{Volume}} 2: {{Short Papers}})}.\hskip 1em
  plus 0.5em minus 0.4em\relax {Melbourne, Australia}: {Association for
  Computational Linguistics}, Jul. 2018, pp. 201--207.

\bibitem{xu2019endtoend}
L.~Xu, Q.~Zhou, K.~Gong, X.~Liang, J.~Tang, and L.~Lin, ``End-to-end
  knowledge-routed relational dialogue system for automatic diagnosis,''
  \emph{arXiv:1901.10623 [cs]}, Mar. 2019.

\bibitem{kawata2017lifestyle}
H.~Kawata, K.~Ookawara, M.~Muta, S.~Masuko, and J.~Hoshino,
  ``\BIBforeignlanguage{en}{Lifestyle agent: The chat-oriented dialogue system
  for lifestyle management},'' in \emph{\BIBforeignlanguage{en}{Entertainment
  {{Computing}} \textendash{} {{ICEC}} 2017}}, ser. Lecture {{Notes}} in
  {{Computer Science}}, N.~Munekata, I.~Kunita, and J.~Hoshino, Eds.\hskip 1em
  plus 0.5em minus 0.4em\relax {Cham}: {Springer International Publishing},
  2017, pp. 396--399.

\bibitem{studer1998knowledge}
R.~Studer, V.~R. Benjamins, and D.~Fensel, ``\BIBforeignlanguage{en}{Knowledge
  engineering: Principles and methods},'' \emph{\BIBforeignlanguage{en}{Data \&
  Knowledge Engineering}}, vol.~25, no.~1, pp. 161--197, Mar. 1998.

\bibitem{goodwin2013automatic}
T.~Goodwin and S.~M. Harabagiu, ``Automatic generation of a qualified medical
  knowledge graph and its usage for retrieving patient cohorts from electronic
  medical records,'' in \emph{2013 {{IEEE Seventh International Conference}} on
  {{Semantic Computing}}}, Sep. 2013, pp. 363--370.

\bibitem{rossanez2020kgen}
A.~Rossanez, J.~C. {dos Reis}, R.~d.~S. Torres, and H.~{de Ribaupierre},
  ``{{KGen}}: A knowledge graph generator from biomedical scientific
  literature,'' \emph{BMC Medical Informatics and Decision Making}, vol.~20,
  no.~4, p. 314, Dec. 2020.

\bibitem{rotmensch2017learning}
M.~Rotmensch, Y.~Halpern, A.~Tlimat, S.~Horng, and D.~Sontag,
  ``\BIBforeignlanguage{en}{Learning a health knowledge graph from electronic
  medical records},'' \emph{\BIBforeignlanguage{en}{Scientific Reports}},
  vol.~7, no.~1, p. 5994, Jul. 2017.

\bibitem{wang2017semantically}
H.~Wang, Q.~Zhang, and J.~Yuan, ``Semantically enhanced medical information
  retrieval system: A tensor factorization based approach,'' \emph{IEEE
  Access}, vol.~5, pp. 7584--7593, 2017.

\bibitem{pan2020medwriter}
Y.~Pan, Q.~Chen, W.~Peng, X.~Wang, B.~Hu, X.~Liu, J.~Chen, and W.~Zhou,
  ``{{MedWriter}}: Knowledge-aware medical text generation,'' in
  \emph{Proceedings of the 28th {{International Conference}} on {{Computational
  Linguistics}}}.\hskip 1em plus 0.5em minus 0.4em\relax {Barcelona, Spain
  (Online)}: {International Committee on Computational Linguistics}, Dec. 2020,
  pp. 2363--2368.

\bibitem{stoica2021intent}
A.~Stoica, T.~Kadar, C.~Lemnaru, R.~Potolea, and M.~D{\^i}n{\c s}oreanu,
  ``\BIBforeignlanguage{en}{Intent detection and slot filling with capsule net
  architectures for a romanian home assistant},''
  \emph{\BIBforeignlanguage{en}{Sensors}}, vol.~21, no.~4, p. 1230, Jan. 2021.

\bibitem{neuraz2018natural}
A.~Neuraz, L.~C. Llanos, A.~Burgun, and S.~Rosset, ``Natural language
  understanding for task oriented dialog in the biomedical domain in a low
  resources context,'' \emph{arXiv:1811.09417 [cs]}, Nov. 2018.

\bibitem{zhang2017bringing}
C.~Zhang, N.~Du, W.~Fan, Y.~Li, C.-T. Lu, and P.~S. Yu, ``Bringing semantic
  structures to user intent detection in online medical queries,'' in
  \emph{2017 {{IEEE International Conference}} on {{Big Data}} ({{Big Data}})},
  Dec. 2017, pp. 1019--1026.

\bibitem{giachos2017exploring}
I.~Giachos, E.~C. Papakitsos, and G.~Chorozoglou,
  ``\BIBforeignlanguage{en}{Exploring natural language understanding in robotic
  interfaces},'' \emph{\BIBforeignlanguage{en}{International Journal of
  Advances in Intelligent Informatics}}, vol.~3, no.~1, pp. 10--19, Mar. 2017.

\bibitem{thomason2019improving}
J.~Thomason, A.~Padmakumar, J.~Sinapov, N.~Walker, Y.~Jiang, H.~Yedidsion,
  J.~Hart, P.~Stone, and R.~J. Mooney, ``Improving grounded natural language
  understanding through human-robot dialog,'' \emph{2019 International
  Conference on Robotics and Automation (ICRA)}, pp. 6934--6941, May 2019.

\bibitem{feder2021causal}
A.~Feder, K.~A. Keith, E.~Manzoor, R.~Pryzant, D.~Sridhar, Z.~{Wood-Doughty},
  J.~Eisenstein, J.~Grimmer, R.~Reichart, M.~E. Roberts, B.~M. Stewart,
  V.~Veitch, and D.~Yang, ``Causal inference in natural language processing:
  Estimation, prediction, interpretation and beyond,'' \emph{arXiv:2109.00725
  [cs]}, Sep. 2021.

\bibitem{zeng2021uncovering}
J.~Zeng, M.~F. Gensheimer, D.~L. Rubin, S.~Athey, and R.~D. Shachter,
  ``Uncovering interpretable potential confounders in electronic medical
  records,'' \emph{medRxiv : the preprint server for health sciences}, 2021.

\bibitem{doan2019extracting}
S.~Doan, E.~W. Yang, S.~S. Tilak, P.~W. Li, D.~S. Zisook, and M.~Torii,
  ``Extracting health-related causality from twitter messages using natural
  language processing,'' \emph{BMC Medical Informatics and Decision Making},
  vol.~19, no.~3, p.~79, Apr. 2019.

\bibitem{nordon2019building}
G.~Nordon, G.~Koren, V.~Shalev, B.~Kimelfeld, U.~Shalit, and K.~Radinsky,
  ``\BIBforeignlanguage{en}{Building causal graphs from medical literature and
  electronic medical records},'' \emph{\BIBforeignlanguage{en}{Proceedings of
  the AAAI Conference on Artificial Intelligence}}, vol.~33, no.~01, pp.
  1102--1109, Jul. 2019.

\bibitem{stiefelhagen2004natural}
R.~Stiefelhagen, C.~Fugen, R.~Gieselmann, H.~Holzapfel, K.~Nickel, and
  A.~Waibel, ``Natural human-robot interaction using speech, head pose and
  gestures,'' in \emph{2004 {{IEEE}}/{{RSJ International Conference}} on
  {{Intelligent Robots}} and {{Systems}} ({{IROS}}) ({{IEEE Cat}}.
  {{No}}.{{04CH37566}})}, vol.~3, Sep. 2004, pp. 2422--2427 vol.3.

\bibitem{finley2018semisupervised}
G.~P. Finley, E.~Edwards, W.~Salloum, A.~Robinson, N.~Sadoughi, N.~Axtmann,
  M.~Korenevsky, M.~Brenndoerfer, M.~Miller, and D.~{Suendermann-Oeft},
  ``\BIBforeignlanguage{en}{Semi-supervised acoustic model retraining for
  medical {{ASR}}},'' in \emph{\BIBforeignlanguage{en}{Speech and
  {{Computer}}}}, ser. Lecture {{Notes}} in {{Computer Science}}, A.~Karpov,
  O.~Jokisch, and R.~Potapova, Eds.\hskip 1em plus 0.5em minus 0.4em\relax
  {Cham}: {Springer International Publishing}, 2018, pp. 177--187.

\bibitem{sas2011optimal}
J.~Sas and T.~Poreba, ``\BIBforeignlanguage{EN}{Optimal acoustic model
  complexity selection in polish medical speech recognition},''
  \emph{\BIBforeignlanguage{EN}{Journal of Medical Informatics \&
  Technologies}}, vol. Vol. 17, 2011.

\bibitem{paulett2009improving}
J.~M. Paulett and C.~P. Langlotz, ``\BIBforeignlanguage{en}{Improving language
  models for radiology speech recognition},''
  \emph{\BIBforeignlanguage{en}{Journal of Biomedical Informatics}}, vol.~42,
  no.~1, pp. 53--58, Feb. 2009.

\bibitem{chiu2018speech}
C.-C. Chiu, A.~Tripathi, K.~Chou, C.~Co, N.~Jaitly, D.~Jaunzeikare, A.~Kannan,
  P.~Nguyen, H.~Sak, A.~Sankar, J.~Tansuwan, N.~Wan, Y.~Wu, and X.~Zhang,
  ``Speech recognition for medical conversations,'' \emph{arXiv:1711.07274 [cs,
  eess, stat]}, Jun. 2018.

\bibitem{edwards2017medical}
E.~Edwards, W.~Salloum, G.~P. Finley, J.~Fone, G.~Cardiff, M.~Miller, and
  D.~{Suendermann-Oeft}, ``\BIBforeignlanguage{en}{Medical speech recognition:
  Reaching parity with humans},'' in \emph{\BIBforeignlanguage{en}{Speech and
  {{Computer}}}}, ser. Lecture {{Notes}} in {{Computer Science}}, A.~Karpov,
  R.~Potapova, and I.~Mporas, Eds.\hskip 1em plus 0.5em minus 0.4em\relax
  {Cham}: {Springer International Publishing}, 2017, pp. 512--524.

\bibitem{he2020doptacotron}
T.~He, W.~Zhao, and L.~Xu, ``{{DOP}}-tacotron: A fast chinese {{TTS}} system
  with local-based attention,'' in \emph{2020 {{Chinese Control And Decision
  Conference}} ({{CCDC}})}, Aug. 2020, pp. 4345--4350.

\bibitem{sugiura2014nonmonologue}
K.~Sugiura, Y.~Shiga, H.~Kawai, T.~Misu, and C.~Hori, ``Non-monologue
  {{HMM}}-based speech synthesis for service robots: {{A}} cloud robotics
  approach,'' in \emph{2014 {{IEEE International Conference}} on {{Robotics}}
  and {{Automation}} ({{ICRA}})}, May 2014, pp. 2237--2242.

\bibitem{akbari2019reconstructing}
H.~Akbari, B.~Khalighinejad, J.~L. Herrero, A.~D. Mehta, and N.~Mesgarani,
  ``\BIBforeignlanguage{en}{Towards reconstructing intelligible speech from the
  human auditory cortex},'' \emph{\BIBforeignlanguage{en}{Scientific Reports}},
  vol.~9, no.~1, p. 874, Jan. 2019.

\bibitem{anumanchipalli2019speech}
G.~K. Anumanchipalli, J.~Chartier, and E.~F. Chang,
  ``\BIBforeignlanguage{en}{Speech synthesis from neural decoding of spoken
  sentences},'' \emph{\BIBforeignlanguage{en}{Nature}}, vol. 568, no. 7753, pp.
  493--498, Apr. 2019.

\bibitem{herff2019generating}
C.~Herff, L.~Diener, M.~Angrick, E.~Mugler, M.~C. Tate, M.~A. Goldrick, D.~J.
  Krusienski, M.~W. Slutzky, and T.~Schultz, ``Generating natural, intelligible
  speech from brain activity in motor, premotor, and inferior frontal
  cortices,'' \emph{Frontiers in Neuroscience}, vol.~13, p. 1267, 2019.

\bibitem{jreige2009vocalid}
C.~Jreige, R.~Patel, and H.~T. Bunnell, ``{{VocaliD}}: Personalizing
  text-to-speech synthesis for individuals with severe speech impairment,'' in
  \emph{Proceedings of the 11th International {{ACM SIGACCESS}} Conference on
  {{Computers}} and Accessibility}, ser. Assets '09.\hskip 1em plus 0.5em minus
  0.4em\relax {New York, NY, USA}: {Association for Computing Machinery}, Oct.
  2009, pp. 259--260.

\bibitem{marge2022spoken}
M.~Marge, C.~{Espy-Wilson}, N.~G. Ward, A.~Alwan, Y.~Artzi, M.~Bansal,
  G.~Blankenship, J.~Chai, H.~Daum{\'e}, D.~Dey, M.~Harper, T.~Howard,
  C.~Kennington, I.~{Kruijff-Korbayov{\'a}}, D.~Manocha, C.~Matuszek, R.~Mead,
  R.~Mooney, R.~K. Moore, M.~Ostendorf, H.~{Pon-Barry}, A.~I. Rudnicky,
  M.~Scheutz, R.~S. Amant, T.~Sun, S.~Tellex, D.~Traum, and Z.~Yu,
  ``\BIBforeignlanguage{en}{Spoken language interaction with robots:
  Recommendations for future research},''
  \emph{\BIBforeignlanguage{en}{Computer Speech \& Language}}, vol.~71, p.
  101255, Jan. 2022.

\bibitem{james2020empathetic}
J.~James, B.~T. Balamurali, C.~I. Watson, and B.~MacDonald,
  ``\BIBforeignlanguage{en}{Empathetic speech synthesis and testing for
  healthcare robots},'' \emph{\BIBforeignlanguage{en}{International Journal of
  Social Robotics}}, Sep. 2020.

\bibitem{li2009expressive}
X.~Li, B.~MacDonald, and C.~I. Watson, ``Expressive facial speech synthesis on
  a robotic platform,'' in \emph{Proceedings of the 2009 {{IEEE}}/{{RSJ}}
  International Conference on {{Intelligent}} Robots and Systems}, ser.
  {{IROS}}'09.\hskip 1em plus 0.5em minus 0.4em\relax {St. Louis, MO, USA}:
  {IEEE Press}, Oct. 2009, pp. 5009--5014.

\bibitem{roehling2006expressive}
S.~Roehling, B.~Macdonald, and C.~Watson, ``Towards expressive speech synthesis
  in english on a robotic platform,'' in \emph{In {{Proceedings}} of the
  {{Australasian International Conference}} on {{Speech Science}} and
  {{Technology}}}, 2006, pp. 130--135.

\bibitem{kuhne2020human}
K.~K{\"u}hne, M.~H. Fischer, and Y.~Zhou, ``The human takes it all: Humanlike
  synthesized voices are perceived as less eerie and more likable. evidence
  from a subjective ratings study,'' \emph{Frontiers in Neurorobotics},
  vol.~14, p. 105, 2020.

\bibitem{jiang2017artificial}
F.~Jiang, Y.~Jiang, H.~Zhi, Y.~Dong, H.~Li, S.~Ma, Y.~Wang, Q.~Dong, H.~Shen,
  and Y.~Wang, ``\BIBforeignlanguage{en}{Artificial intelligence in healthcare:
  Past, present and future},'' \emph{\BIBforeignlanguage{en}{Stroke and
  Vascular Neurology}}, vol.~2, no.~4, Dec. 2017.

\bibitem{dew2018development}
K.~N. Dew, A.~M. Turner, Y.~K. Choi, A.~Bosold, and K.~Kirchhoff,
  ``\BIBforeignlanguage{en}{Development of machine translation technology for
  assisting health communication: {{A}} systematic review},''
  \emph{\BIBforeignlanguage{en}{Journal of Biomedical Informatics}}, vol.~85,
  pp. 56--67, Sep. 2018.

\bibitem{randhawa2013using}
G.~Randhawa, M.~Ferreyra, R.~Ahmed, O.~Ezzat, and K.~Pottie, ``Using machine
  translation in clinical practice,'' \emph{Canadian Family Physician},
  vol.~59, no.~4, pp. 382--383, Apr. 2013.

\bibitem{goss2019clinician}
F.~Goss, S.~Blackley, C.~Ortega, L.~Kowalski, A.~Landman, C.~Lin, M.~Meteer,
  S.~Bakes, S.~Gradwohl, D.~Bates, and Z.~Li, ``A clinician survey of using
  speech recognition for clinical documentation in the electronic health
  record,'' \emph{International Journal of Medical Informatics}, vol. 130, Jul.
  2019.

\bibitem{saxena2018provider}
K.~Saxena, R.~Diamond, R.~F. Conant, T.~H. Mitchell, i.~G. Gallopyn, and K.~E.
  Yakimow, ``Provider adoption of speech recognition and its impact on
  satisfaction, documentation quality, efficiency, and cost in an inpatient
  {{EHR}},'' \emph{AMIA Summits on Translational Science Proceedings}, vol.
  2018, pp. 186--195, May 2018.

\bibitem{zhao2009speechrecognition}
Y.~Zhao, ``Speech-recognition technology in health care and special-needs
  assistance [life sciences],'' \emph{IEEE Signal Processing Magazine},
  vol.~26, no.~3, pp. 87--90, May 2009.

\bibitem{goodwin2016medical}
T.~R. Goodwin and S.~M. Harabagiu, ``\BIBforeignlanguage{eng}{Medical question
  answering for clinical decision support},''
  \emph{\BIBforeignlanguage{eng}{Proceedings of the ... ACM International
  Conference on Information \& Knowledge Management. ACM International
  Conference on Information and Knowledge Management}}, vol. 2016, pp.
  297--306, Oct. 2016.

\bibitem{xu2021external}
G.~Xu, W.~Rong, Y.~Wang, Y.~Ouyang, and Z.~Xiong,
  ``\BIBforeignlanguage{en}{External features enriched model for biomedical
  question answering},'' \emph{\BIBforeignlanguage{en}{BMC Bioinformatics}},
  vol.~22, no.~1, p. 272, May 2021.

\bibitem{shi2019family}
X.~Shi, D.~Jiang, Y.~Huang, X.~Wang, Q.~Chen, J.~Yan, and B.~Tang, ``Family
  history information extraction via deep joint learning,'' \emph{BMC Medical
  Informatics and Decision Making}, vol.~19, no. Suppl 10, Dec. 2019.

\bibitem{gupta2018automatic}
A.~Gupta, I.~Banerjee, and D.~L. Rubin, ``\BIBforeignlanguage{en}{Automatic
  information extraction from unstructured mammography reports using
  distributed semantics},'' \emph{\BIBforeignlanguage{en}{Journal of Biomedical
  Informatics}}, vol.~78, pp. 78--86, Feb. 2018.

\bibitem{yang2019information}
J.~Yang, Y.~Liu, M.~Qian, C.~Guan, and X.~Yuan,
  ``\BIBforeignlanguage{en}{Information extraction from electronic medical
  records using multitask recurrent neural network with contextual word
  embedding},'' \emph{\BIBforeignlanguage{en}{Applied Sciences}}, vol.~9,
  no.~18, p. 3658, Jan. 2019.

\bibitem{zheng2018automated}
S.~Zheng, S.~K. Jabbour, S.~E. O'Reilly, J.~J. Lu, L.~Dong, L.~Ding, Y.~Xiao,
  N.~Yue, F.~Wang, and W.~Zou, ``Automated information extraction on treatment
  and prognosis for {{Non}}\textendash{{Small}} cell lung cancer radiotherapy
  patients: Clinical study,'' \emph{JMIR Medical Informatics}, vol.~6, no.~1,
  p.~e8, Feb. 2018.

\bibitem{liang2019evaluation}
H.~Liang, B.~Y. Tsui, H.~Ni, C.~C.~S. Valentim, S.~L. Baxter, G.~Liu, W.~Cai,
  D.~S. Kermany, X.~Sun, J.~Chen, L.~He, J.~Zhu, P.~Tian, H.~Shao, L.~Zheng,
  R.~Hou, S.~Hewett, G.~Li, P.~Liang, X.~Zang, Z.~Zhang, L.~Pan, H.~Cai,
  R.~Ling, S.~Li, Y.~Cui, S.~Tang, H.~Ye, X.~Huang, W.~He, W.~Liang, Q.~Zhang,
  J.~Jiang, W.~Yu, J.~Gao, W.~Ou, Y.~Deng, Q.~Hou, B.~Wang, C.~Yao, Y.~Liang,
  S.~Zhang, Y.~Duan, R.~Zhang, S.~Gibson, C.~L. Zhang, O.~Li, E.~D. Zhang,
  G.~Karin, N.~Nguyen, X.~Wu, C.~Wen, J.~Xu, W.~Xu, B.~Wang, W.~Wang, J.~Li,
  B.~Pizzato, C.~Bao, D.~Xiang, W.~He, S.~He, Y.~Zhou, W.~Haw, M.~Goldbaum,
  A.~Tremoulet, C.-N. Hsu, H.~Carter, L.~Zhu, K.~Zhang, and H.~Xia,
  ``\BIBforeignlanguage{en}{Evaluation and accurate diagnoses of pediatric
  diseases using artificial intelligence},''
  \emph{\BIBforeignlanguage{en}{Nature Medicine}}, vol.~25, no.~3, pp.
  433--438, Mar. 2019.

\bibitem{harkema2011developing}
H.~Harkema, W.~W. Chapman, M.~Saul, E.~S. Dellon, R.~E. Schoen, and
  A.~Mehrotra, ``\BIBforeignlanguage{eng}{Developing a natural language
  processing application for measuring the quality of colonoscopy
  procedures},'' \emph{\BIBforeignlanguage{eng}{Journal of the American Medical
  Informatics Association: JAMIA}}, vol. 18 Suppl 1, pp. i150--156, Dec. 2011.

\bibitem{mehrotra2012applying}
A.~Mehrotra, E.~S. Dellon, R.~E. Schoen, M.~Saul, F.~Bishehsari, C.~Farmer, and
  H.~Harkema, ``\BIBforeignlanguage{English}{Applying a natural language
  processing tool to electronic health records to assess performance on
  colonoscopy quality measures},''
  \emph{\BIBforeignlanguage{English}{Gastrointestinal Endoscopy}}, vol.~75,
  no.~6, pp. 1233--1239.e14, Jun. 2012.

\bibitem{wunnava2019adverse}
S.~Wunnava, X.~Qin, T.~Kakar, C.~Sen, E.~A. Rundensteiner, and X.~Kong,
  ``\BIBforeignlanguage{eng}{Adverse drug event detection from electronic
  health records using hierarchical recurrent neural networks with dual-level
  embedding},'' \emph{\BIBforeignlanguage{eng}{Drug Safety}}, vol.~42, no.~1,
  pp. 113--122, Jan. 2019.

\bibitem{jackson2017natural}
R.~G. Jackson, R.~Patel, N.~Jayatilleke, A.~Kolliakou, M.~Ball, G.~Gorrell,
  A.~Roberts, R.~J. Dobson, and R.~Stewart, ``\BIBforeignlanguage{en}{Natural
  language processing to extract symptoms of severe mental illness from
  clinical text: The clinical record interactive search comprehensive data
  extraction ({{CRIS}}-{{CODE}}) project},'' \emph{\BIBforeignlanguage{en}{BMJ
  Open}}, vol.~7, no.~1, p. e012012, Jan. 2017.

\bibitem{luo2021early}
J.~Luo, L.~Lan, D.~Yang, S.~Huang, M.~Li, J.~Yin, J.~Xiao, and X.~Zhou,
  ``\BIBforeignlanguage{en}{Early prediction of organ failures in patients with
  acute pancreatitis using text mining},''
  \emph{\BIBforeignlanguage{en}{Scientific Programming}}, vol. 2021, p.
  e6683942, May 2021.

\bibitem{wang2021inferbert}
X.~Wang, X.~Xu, W.~Tong, R.~Roberts, and Z.~Liu, ``{{InferBERT}}: {{A}}
  transformer-based causal inference framework for enhancing
  pharmacovigilance,'' \emph{Frontiers in Artificial Intelligence}, vol.~4,
  p.~67, 2021.

\bibitem{wang2019transparent}
S.~V. Wang, O.~V. Patterson, J.~J. Gagne, J.~S. Brown, R.~Ball, P.~Jonsson,
  A.~Wright, L.~Zhou, W.~Goettsch, and A.~Bate,
  ``\BIBforeignlanguage{en}{Transparent reporting on research using
  unstructured electronic health record data to generate `real world' evidence
  of comparative effectiveness and safety},''
  \emph{\BIBforeignlanguage{en}{Drug Safety}}, vol.~42, no.~11, pp. 1297--1309,
  Nov. 2019.

\bibitem{lee2019information}
A.~Lee, B.~E. Alving, M.~B. Horup, and L.~Thrysoee,
  ``\BIBforeignlanguage{en}{Information retrieval as a part of evidence-based
  practice: Retrieval skills, behavior and needs among nurses at a large
  university hospital:},'' \emph{\BIBforeignlanguage{en}{Nordic Journal of
  Nursing Research}}, Aug. 2019.

\bibitem{patrick2004evidencebased}
T.~B. Patrick, G.~Demiris, L.~C. Folk, D.~E. Moxley, J.~A. Mitchell, and
  D.~Tao, ``Evidence-based retrieval in evidence-based medicine,''
  \emph{Journal of the Medical Library Association}, vol.~92, no.~2, pp.
  196--199, Apr. 2004.

\bibitem{ford1999information}
N.~Ford, D.~Miller, A.~Booth, A.~O'rourke, J.~Ralph, and E.~Turnock,
  ``Information retrieval for evidence-based decision making,'' \emph{JOURNAL
  OF DOCUMENTATION}, vol.~55, Oct. 1999.

\bibitem{sterling2019prediction}
N.~W. Sterling, R.~E. Patzer, M.~Di, and J.~D. Schrager,
  ``\BIBforeignlanguage{en}{Prediction of emergency department patient
  disposition based on natural language processing of triage notes},''
  \emph{\BIBforeignlanguage{en}{International Journal of Medical Informatics}},
  vol. 129, pp. 184--188, Sep. 2019.

\bibitem{tahayori2021advanced}
B.~Tahayori, N.~{Chini-Foroush}, and H.~Akhlaghi,
  ``\BIBforeignlanguage{en}{Advanced natural language processing technique to
  predict patient disposition based on emergency triage notes},''
  \emph{\BIBforeignlanguage{en}{Emergency Medicine Australasia}}, vol.~33,
  no.~3, pp. 480--484, 2021.

\bibitem{sezgin2020readiness}
E.~Sezgin, Y.~Huang, U.~Ramtekkar, and S.~Lin,
  ``\BIBforeignlanguage{en}{Readiness for voice assistants to support
  healthcare delivery during a health crisis and pandemic},''
  \emph{\BIBforeignlanguage{en}{npj Digital Medicine}}, vol.~3, no.~1, pp.
  1--4, Sep. 2020.

\bibitem{spanig2019virtual}
S.~Sp{\"a}nig, A.~{Emberger-Klein}, J.-P. Sowa, A.~Canbay, K.~Menrad, and
  D.~Heider, ``The virtual doctor: An interactive artificial intelligence based
  on deep learning for non-invasive prediction of diabetes,'' \emph{Artificial
  Intelligence in Medicine}, vol. 100, p. 101706, Sep. 2019.

\bibitem{gandhi2019intellidoctor}
M.~Gandhi, V.~K. Singh, and V.~Kumar, ``{{IntelliDoctor}} - {{AI}} based
  medical assistant,'' in \emph{2019 {{Fifth International Conference}} on
  {{Science Technology Engineering}} and {{Mathematics}} ({{ICONSTEM}})},
  vol.~1, Mar. 2019, pp. 162--168.

\bibitem{agustin2019voice}
E.~I. Agustin, R.~T. Yunardi, and A.~A. Firdaus,
  ``\BIBforeignlanguage{en-US}{Voice recognition system for controlling
  electrical appliances in smart hospital room},''
  \emph{\BIBforeignlanguage{en-US}{TELKOMNIKA (Telecommunication Computing
  Electronics and Control)}}, vol.~17, no.~2, pp. 965--972, Apr. 2019.

\bibitem{ismail2020development}
A.~Ismail, S.~Abdlerazek, and I.~M. {El-Henawy},
  ``\BIBforeignlanguage{en}{Development of smart healthcare system based on
  speech recognition using support vector machine and dynamic time warping},''
  \emph{\BIBforeignlanguage{en}{Sustainability}}, vol.~12, no.~6, p. 2403, Jan.
  2020.

\bibitem{grasse2021speech}
L.~Grasse, S.~J. Boutros, and M.~S. Tata, ``Speech interaction to control a
  hands-free delivery robot for high-risk health care scenarios,''
  \emph{Frontiers in Robotics and AI}, vol.~8, p.~40, 2021.

\bibitem{holland2021service}
J.~Holland, L.~Kingston, C.~McCarthy, E.~Armstrong, P.~O'Dwyer, F.~Merz, and
  M.~McConnell, ``\BIBforeignlanguage{en}{Service robots in the healthcare
  sector},'' \emph{\BIBforeignlanguage{en}{Robotics}}, vol.~10, no.~1, p.~47,
  Mar. 2021.

\bibitem{lineback2021prediction}
C.~M. Lineback, R.~Garg, E.~Oh, A.~M. Naidech, J.~L. Holl, and S.~Prabhakaran,
  ``Prediction of 30-day readmission after stroke using machine learning and
  natural language processing,'' \emph{Frontiers in Neurology}, vol.~12, p.
  1069, 2021.

\bibitem{rajkomar2018scalable}
A.~Rajkomar, E.~Oren, K.~Chen, A.~M. Dai, N.~Hajaj, M.~Hardt, P.~J. Liu,
  X.~Liu, J.~Marcus, M.~Sun, P.~Sundberg, H.~Yee, K.~Zhang, Y.~Zhang,
  G.~Flores, G.~E. Duggan, J.~Irvine, Q.~Le, K.~Litsch, A.~Mossin, J.~Tansuwan,
  D.~Wang, J.~Wexler, J.~Wilson, D.~Ludwig, S.~L. Volchenboum, K.~Chou,
  M.~Pearson, S.~Madabushi, N.~H. Shah, A.~J. Butte, M.~D. Howell, C.~Cui,
  G.~S. Corrado, and J.~Dean, ``\BIBforeignlanguage{en}{Scalable and accurate
  deep learning with electronic health records},''
  \emph{\BIBforeignlanguage{en}{npj Digital Medicine}}, vol.~1, no.~1, p.~18,
  Dec. 2018.

\bibitem{rumshisky2016predicting}
A.~Rumshisky, M.~Ghassemi, T.~Naumann, P.~Szolovits, V.~M. Castro, T.~H. McCoy,
  and R.~H. Perlis, ``Predicting early psychiatric readmission with natural
  language processing of narrative discharge summaries,'' \emph{Translational
  Psychiatry}, vol.~6, no.~10, p. e921, Oct. 2016.

\bibitem{alfarghaly2021automated}
O.~Alfarghaly, R.~Khaled, A.~Elkorany, M.~Helal, and A.~Fahmy,
  ``\BIBforeignlanguage{en}{Automated radiology report generation using
  conditioned transformers},'' \emph{\BIBforeignlanguage{en}{Informatics in
  Medicine Unlocked}}, vol.~24, p. 100557, Jan. 2021.

\bibitem{chintagunta2021medically}
B.~Chintagunta, N.~Katariya, X.~Amatriain, and A.~Kannan, ``Medically aware
  {{GPT}}-3 as a data generator for medical dialogue summarization,'' in
  \emph{Proceedings of the {{Second Workshop}} on {{Natural Language
  Processing}} for {{Medical Conversations}}}.\hskip 1em plus 0.5em minus
  0.4em\relax {Online}: {Association for Computational Linguistics}, Jun. 2021,
  pp. 66--76.

\bibitem{ibrahim2013ontologydriven}
A.~M. Ibrahim, ``\BIBforeignlanguage{en}{Ontology-driven information retrieval
  for healthcare information system : {{A}} case study},''
  \emph{\BIBforeignlanguage{en}{International Journal of Network Security \&
  Its Applications}}, vol.~5, no.~1, pp. 61--69, Jan. 2013.

\bibitem{khanbhai2021applying}
M.~Khanbhai, P.~Anyadi, J.~Symons, K.~Flott, A.~Darzi, and E.~Mayer,
  ``\BIBforeignlanguage{en}{Applying natural language processing and machine
  learning techniques to patient experience feedback: A systematic review},''
  \emph{\BIBforeignlanguage{en}{BMJ Health \& Care Informatics}}, vol.~28,
  no.~1, p. e100262, Mar. 2021.

\bibitem{nawab2020natural}
K.~Nawab, G.~Ramsey, and R.~Schreiber, ``\BIBforeignlanguage{eng}{Natural
  language processing to extract meaningful information from patient experience
  feedback},'' \emph{\BIBforeignlanguage{eng}{Applied Clinical Informatics}},
  vol.~11, no.~2, pp. 242--252, Mar. 2020.

\bibitem{doing-harris2017understanding}
K.~{Doing-Harris}, D.~L. Mowery, C.~Daniels, W.~W. Chapman, and M.~Conway,
  ``Understanding patient satisfaction with received healthcare services: {{A}}
  natural language processing approach,'' \emph{AMIA Annual Symposium
  Proceedings}, vol. 2016, pp. 524--533, Feb. 2017.

\bibitem{rodger2013pregnant}
D.~Rodger, A.~Skuse, M.~Wilmore, S.~Humphreys, J.~Dalton, M.~Flabouris, V.~L.
  Clifton, D.~Rodger, A.~Skuse, M.~Wilmore, S.~Humphreys, J.~Dalton,
  M.~Flabouris, and V.~L. Clifton, ``\BIBforeignlanguage{en}{Pregnant women's
  use of information and communications technologies to access
  pregnancy-related health information in {{South Australia}}},''
  \emph{\BIBforeignlanguage{en}{Australian Journal of Primary Health}},
  vol.~19, no.~4, pp. 308--312, Dec. 2013.

\bibitem{zhou2018new}
B.~Zhou, K.~Wu, P.~Lv, J.~Wang, G.~Chen, B.~Ji, and S.~Liu, ``A new remote
  health-care system based on moving robot intended for the elderly at home,''
  \emph{Journal of Healthcare Engineering}, vol. 2018, p. 4949863, Feb. 2018.

\bibitem{rani2017voice}
P.~J. Rani, J.~Bakthakumar, B.~P. Kumaar, U.~P. Kumaar, and S.~Kumar, ``Voice
  controlled home automation system using natural language processing ({{NLP}})
  and internet of things ({{IoT}}),'' in \emph{2017 {{Third International
  Conference}} on {{Science Technology Engineering Management}}
  ({{ICONSTEM}})}, Mar. 2017, pp. 368--373.

\bibitem{tapus2007socially}
A.~Tapus, M.~J. Mataric, and B.~Scassellati, ``Socially assistive robotics,''
  \emph{IEEE Robotics Automation Magazine}, vol.~14, no.~1, pp. 35--42, Mar.
  2007.

\bibitem{mavridis2015review}
N.~Mavridis, ``\BIBforeignlanguage{en}{A review of verbal and non-verbal
  human\textendash robot interactive communication},''
  \emph{\BIBforeignlanguage{en}{Robotics and Autonomous Systems}}, vol.~63, pp.
  22--35, Jan. 2015.

\bibitem{green2021automatic}
J.~R. Green, R.~L. MacDonald, P.-P. Jiang, J.~Cattiau, R.~Heywood, R.~Cave,
  K.~Seaver, M.~A. Ladewig, J.~Tobin, M.~P. Brenner, P.~C. Nelson, and
  K.~Tomanek, ``\BIBforeignlanguage{en}{Automatic speech recognition of
  disordered speech: Personalized models outperforming human listeners on short
  phrases},'' in \emph{\BIBforeignlanguage{en}{Interspeech 2021}}.\hskip 1em
  plus 0.5em minus 0.4em\relax {ISCA}, Aug. 2021, pp. 4778--4782.

\bibitem{hux2017comprehension}
K.~Hux, K.~{Knollman-Porter}, J.~Brown, and S.~E. Wallace,
  ``\BIBforeignlanguage{en}{Comprehension of synthetic speech and digitized
  natural speech by adults with aphasia},''
  \emph{\BIBforeignlanguage{en}{Journal of Communication Disorders}}, vol.~69,
  pp. 15--26, Sep. 2017.

\bibitem{hux2020effect}
K.~Hux, J.~A. Brown, S.~Wallace, K.~{Knollman-Porter}, A.~Saylor, and E.~Lapp,
  ``\BIBforeignlanguage{eng}{Effect of text-to-speech rate on reading
  comprehension by adults with aphasia},''
  \emph{\BIBforeignlanguage{eng}{American Journal of Speech-Language
  Pathology}}, vol.~29, no.~1, pp. 168--184, Jul. 2020.

\bibitem{cassidy2016expressive}
S.~Cassidy, B.~Stenger, L.~Van~Dongen, K.~Yanagisawa, R.~Anderson, V.~Wan,
  S.~{Baron-Cohen}, and R.~Cipolla, ``Expressive visual text-to-speech as an
  assistive technology for individuals with autism spectrum conditions,''
  \emph{Computer Vision and Image Understanding}, vol. 148, pp. 193--200, Jul.
  2016.

\bibitem{repova2020texttospeech}
B.~Repova, M.~Zabrodsky, J.~Plzak, D.~Kalfert, J.~Matousek, and J.~Betka,
  ``\BIBforeignlanguage{eng}{Text-to-speech synthesis as an alternative
  communication means after total laryngectomy},''
  \emph{\BIBforeignlanguage{eng}{Biomedical Papers of the Medical Faculty of
  the University Palacky, Olomouc, Czechoslovakia}}, Apr. 2020.

\bibitem{lyall2016smartphone}
F.~C. Lyall, P.~J. Clamp, and D.~Hajioff, ``\BIBforeignlanguage{eng}{Smartphone
  speech-to-text applications for communication with profoundly deaf
  patients},'' \emph{\BIBforeignlanguage{eng}{The Journal of Laryngology and
  Otology}}, vol. 130, no.~1, pp. 104--106, Jan. 2016.

\bibitem{nittrouer2018speech}
S.~Nittrouer, L.~M. Krieg, and J.~H. Lowenstein, ``Speech recognition in noise
  by children with and without dyslexia: How is it related to reading?''
  \emph{Research in developmental disabilities}, vol.~77, pp. 98--113, Jun.
  2018.

\bibitem{riano2019ten}
D.~Ria{\~n}o, M.~Peleg, and A.~{ten Teije}, ``\BIBforeignlanguage{en}{Ten years
  of knowledge representation for health care (2009\textendash 2018): Topics,
  trends, and challenges},'' \emph{\BIBforeignlanguage{en}{Artificial
  Intelligence in Medicine}}, vol. 100, p. 101713, Sep. 2019.

\bibitem{bao2020hhh}
Q.~Bao, L.~Ni, and J.~Liu, ``{{HHH}}: An online medical chatbot system based on
  knowledge graph and hierarchical bi-directional attention,''
  \emph{Proceedings of the Australasian Computer Science Week Multiconference},
  pp. 1--10, Feb. 2020.

\bibitem{xie2018mobilebased}
W.~Xie, R.~Ding, J.~Yan, and Y.~Qu, ``\BIBforeignlanguage{en}{A mobile-based
  question-answering and early warning system for assisting diabetes
  management},'' \emph{\BIBforeignlanguage{en}{Wireless Communications and
  Mobile Computing}}, vol. 2018, p. e9163160, Jun. 2018.

\bibitem{peters2016translating}
P.~Peters, Y.~Qian, and J.~Ding, ``\BIBforeignlanguage{en}{Translating medical
  terminology and bilingual terminography},''
  \emph{\BIBforeignlanguage{en}{Lexicography: Journal of ASIALEX}}, vol.~3,
  no.~2, pp. 99--113, 2016.

\bibitem{renato2018machine}
A.~Renato, J.~Casta{\~n}o, M.~d. P.~A. Williams, H.~Berinsky, M.~L. Gambarte,
  H.~Park, D.~{P{\'e}rez-Rey}, C.~Otero, and D.~Luna, ``A machine translation
  approach for medical terms,'' in \emph{{{HEALTHINF}}}, 2018.

\bibitem{leo2019ontologybased}
J.~Leo, G.~Kurdi, N.~Matentzoglu, B.~Parsia, U.~Sattler, S.~Forge, G.~Donato,
  and W.~Dowling, ``\BIBforeignlanguage{en}{Ontology-based generation of
  medical, multi-term {{MCQs}}},'' \emph{\BIBforeignlanguage{en}{International
  Journal of Artificial Intelligence in Education}}, vol.~29, no.~2, pp.
  145--188, May 2019.

\bibitem{nhs_2013}
\BIBentryALTinterwordspacing
NHS, ``Nhs population screening explained,'' \emph{nhs population screening
  explained}, Feb 2013. [Online]. Available:
  \url{https://www.gov.uk/guidance/nhs-population-screening-explained}
\BIBentrySTDinterwordspacing

\bibitem{m2015state}
P.~M, G.~M, {Newton-DameRemle}, T.~E, P.~E, M.~H, and G.~N,
  ``\BIBforeignlanguage{en}{The state of population health surveillance using
  electronic health records: {{A}} narrative review},''
  \emph{\BIBforeignlanguage{en}{Population Health Management}}, Jun. 2015.

\bibitem{georgiou2015extracting}
D.~Georgiou, A.~MacFarlane, and T.~{Russell-Rose}, ``Extracting sentiment from
  healthcare survey data: An evaluation of sentiment analysis tools,'' in
  \emph{2015 {{Science}} and {{Information Conference}} ({{SAI}})}, Jul. 2015,
  pp. 352--361.

\bibitem{ozturk2020exploring}
H.~{\"O}zt{\"u}rk, A.~{\"O}zg{\"u}r, P.~Schwaller, T.~Laino, and E.~Ozkirimli,
  ``\BIBforeignlanguage{en}{Exploring chemical space using natural language
  processing methodologies for drug discovery},''
  \emph{\BIBforeignlanguage{en}{Drug Discovery Today}}, vol.~25, no.~4, pp.
  689--705, Apr. 2020.

\bibitem{lake2019artificial}
F.~Lake, ``Artificial intelligence in drug discovery: What is new, and what is
  next?'' \emph{Future Drug Discovery}, vol.~1, no.~2, p. FDD19, Oct. 2019.

\bibitem{pham2021deep}
T.-H. Pham, Y.~Qiu, J.~Zeng, L.~Xie, and P.~Zhang, ``\BIBforeignlanguage{en}{A
  deep learning framework for high-throughput mechanism-driven phenotype
  compound screening and its application to {{COVID}}-19 drug repurposing},''
  \emph{\BIBforeignlanguage{en}{Nature Machine Intelligence}}, vol.~3, no.~3,
  pp. 247--257, Mar. 2021.

\bibitem{schwartz2013smifp}
J.~Schwartz, M.~Awale, and J.-L. Reymond, ``{{SMIfp}} ({{SMILES}} fingerprint)
  chemical space for virtual screening and visualization of large databases of
  organic molecules,'' \emph{Journal of Chemical Information and Modeling},
  vol.~53, no.~8, pp. 1979--1989, Aug. 2013.

\bibitem{zhang2021prediction}
F.~Zhang, B.~Sun, X.~Diao, W.~Zhao, and T.~Shu, ``Prediction of adverse drug
  reactions based on knowledge graph embedding,'' \emph{BMC Medical Informatics
  and Decision Making}, vol.~21, no.~1, p.~38, Feb. 2021.

\bibitem{bouhedjar2020natural}
K.~Bouhedjar, A.~Boukelia, A.~K. Nacereddine, A.~Boucheham, A.~Belaidi, and
  A.~Djerourou, ``\BIBforeignlanguage{en}{A natural language processing
  approach based on embedding deep learning from heterogeneous compounds for
  quantitative structure\textendash activity relationship modeling},''
  \emph{\BIBforeignlanguage{en}{Chemical Biology \& Drug Design}}, vol.~96,
  no.~3, pp. 961--972, 2020.

\bibitem{jeon2019fp2vec}
W.~Jeon and D.~Kim, ``{{FP2VEC}}: A new molecular featurizer for learning
  molecular properties,'' \emph{Bioinformatics}, vol.~35, no.~23, pp.
  4979--4985, Dec. 2019.

\bibitem{chen2019clinical}
L.~Chen, Y.~Gu, X.~Ji, C.~Lou, Z.~Sun, H.~Li, Y.~Gao, and Y.~Huang, ``Clinical
  trial cohort selection based on multi-level rule-based natural language
  processing system,'' \emph{Journal of the American Medical Informatics
  Association : JAMIA}, vol.~26, no.~11, pp. 1218--1226, Jul. 2019.

\bibitem{harrer2019artificial}
S.~Harrer, P.~Shah, B.~Antony, and J.~Hu, ``\BIBforeignlanguage{en}{Artificial
  intelligence for clinical trial design},''
  \emph{\BIBforeignlanguage{en}{Trends in Pharmacological Sciences}}, vol.~40,
  no.~8, pp. 577--591, Aug. 2019.

\bibitem{tissot2020natural}
H.~Tissot, F.~Asselbergs, A.~Shah, D.~Brealey, S.~Harris, R.~Agbakoba,
  A.~Folarin, L.~Romao, L.~Roguski, and R.~Dobson, ``Natural language
  processing for mimicking clinical trial recruitment in critical care: {{A}}
  semi-automated simulation based on the {{LeoPARDS}} trial,'' \emph{IEEE
  Journal of Biomedical and Health Informatics}, vol.~PP, pp. 1--1, Mar. 2020.

\bibitem{chen2020trends}
X.~Chen, H.~Xie, G.~Cheng, L.~K.~M. Poon, M.~Leng, and F.~L. Wang,
  ``\BIBforeignlanguage{en}{Trends and features of the applications of natural
  language processing techniques for clinical trials text analysis},''
  \emph{\BIBforeignlanguage{en}{Applied Sciences}}, vol.~10, no.~6, p. 2157,
  Jan. 2020.

\bibitem{ventola2018big}
C.~L. Ventola, ``Big data and pharmacovigilance: Data mining for adverse drug
  events and interactions,'' \emph{Pharmacy and Therapeutics}, vol.~43, no.~6,
  pp. 340--351, Jun. 2018.

\bibitem{wang2009active}
X.~Wang, G.~Hripcsak, M.~Markatou, and C.~Friedman, ``Active computerized
  pharmacovigilance using natural language processing, statistics, and
  electronic health records: {{A}} feasibility study,'' \emph{Journal of the
  American Medical Informatics Association : JAMIA}, vol.~16, no.~3, pp.
  328--337, 2009.

\bibitem{liu2019drug}
F.~Liu, A.~Jagannatha, and H.~Yu, ``\BIBforeignlanguage{en}{Towards drug safety
  surveillance and pharmacovigilance: Current progress in detecting medication
  and adverse drug events from electronic health records},''
  \emph{\BIBforeignlanguage{en}{Drug Safety}}, vol.~42, no.~1, pp. 95--97, Jan.
  2019.

\bibitem{zhou2022interpretable}
B.~Zhou, G.~Yang, Z.~Shi, and S.~Ma, ``Interpretable {{Temporal Attention
  Network}} for {{COVID-19}} forecasting,'' \emph{Applied Soft Computing}, vol.
  120, p. 108691, May 2022.

\bibitem{zheng2020predicting}
N.~Zheng, S.~Du, J.~Wang, H.~Zhang, W.~Cui, Z.~Kang, T.~Yang, B.~Lou, Y.~Chi,
  H.~Long, M.~Ma, Q.~Yuan, S.~Zhang, D.~Zhang, F.~Ye, and J.~Xin, ``Predicting
  {{COVID-19}} in china using hybrid {{AI}} model,'' \emph{IEEE Transactions on
  Cybernetics}, vol.~50, no.~7, pp. 2891--2904, Jul. 2020.

\bibitem{chen2021artificial}
Q.~Chen, R.~Leaman, A.~Allot, L.~Luo, C.-H. Wei, S.~Yan, and Z.~Lu,
  ``Artificial intelligence in action: Addressing the {{COVID-19}} pandemic
  with natural language processing,'' \emph{Annual Review of Biomedical Data
  Science}, vol.~4, no.~1, pp. 313--339, 2021.

\bibitem{chapman2020natural}
A.~Chapman, K.~Peterson, A.~Turano, T.~Box, K.~Wallace, and M.~Jones, ``A
  natural language processing system for national {{COVID-19}} surveillance in
  the {{US}} department of veterans affairs,'' in \emph{Proceedings of the 1st
  {{Workshop}} on {{NLP}} for {{COVID-19}} at {{ACL}} 2020}.\hskip 1em plus
  0.5em minus 0.4em\relax {Online}: {Association for Computational
  Linguistics}, 2020.

\bibitem{cury2021natural}
R.~C. Cury, I.~Megyeri, T.~Lindsey, R.~Macedo, J.~Batlle, S.~Kim, B.~Baker,
  R.~Harris, and R.~H. Clark, ``Natural language processing and machine
  learning for detection of respiratory illness by chest {{CT}} imaging and
  tracking of {{COVID-19}} pandemic in the united states,'' \emph{Radiology:
  Cardiothoracic Imaging}, vol.~3, no.~1, p. e200596, Feb. 2021.

\bibitem{decaprio2020building}
D.~DeCaprio, J.~Gartner, C.~J. McCall, T.~Burgess, K.~Garcia, S.~Kothari, and
  S.~Sayed, ``Building a {{COVID-19}} vulnerability index,'' \emph{Journal of
  Medical Artificial Intelligence}, vol.~3, no.~0, Dec. 2020.

\bibitem{meystre2021natural}
S.~M. Meystre, P.~M. Heider, Y.~Kim, M.~Davis, J.~Obeid, J.~Madory, and A.~V.
  Alekseyenko, ``Natural language processing enabling {{COVID-19}} predictive
  analytics to support data-driven patient advising and pooled testing,''
  \emph{Journal of the American Medical Informatics Association: JAMIA},
  vol.~29, no.~1, pp. 12--21, Dec. 2021.

\bibitem{wang2022novel}
L.~Wang, L.~Jiang, D.~Pan, Q.~Wang, Z.~Yin, Z.~Kang, H.~Tian, X.~Geng, J.~Shao,
  W.~Pan, J.~Yin, L.~Fang, Y.~Wang, W.~Zhang, Z.~Li, J.~Zheng, W.~Hu, Y.~Pan,
  D.~Yu, S.~Guo, W.~Lu, Q.~Li, Y.~Zhou, and H.~Xu, ``Novel approach by natural
  language processing for {{COVID-19}} knowledge discovery,'' \emph{Biomedical
  Journal}, Apr. 2022.

\bibitem{keshavarziarshadi2020artificial}
A.~Keshavarzi~Arshadi, J.~Webb, M.~Salem, E.~Cruz, S.~{Calad-Thomson},
  N.~Ghadirian, J.~Collins, E.~{Diez-Cecilia}, B.~Kelly, H.~Goodarzi, and J.~S.
  Yuan, ``Artificial {{Intelligence}} for {{COVID-19 Drug Discovery}} and
  {{Vaccine Development}},'' \emph{Frontiers in Artificial Intelligence},
  vol.~3, 2020.

\bibitem{liu2021aibased}
Z.~Liu, R.~A. Roberts, M.~{Lal-Nag}, X.~Chen, R.~Huang, and W.~Tong,
  ``{{AI-based}} language models powering drug discovery and development,''
  \emph{Drug Discovery Today}, vol.~26, no.~11, pp. 2593--2607, Nov. 2021.

\bibitem{who202010}
WHO, ``10 global health issues to track in 2021,''
  https://www.who.int/news-room/spotlight/10-global-health-issues-to-track-in-2021,
  2020.

\bibitem{dai2021deepa}
H.-J. Dai, C.-H. Su, Y.-Q. Lee, Y.-C. Zhang, C.-K. Wang, C.-J. Kuo, and C.-S.
  Wu, ``Deep learning-based natural language processing for screening
  psychiatric patients,'' \emph{Frontiers in Psychiatry}, vol.~11, 2021.

\bibitem{desouza2021natural}
D.~D. DeSouza, J.~Robin, M.~Gumus, and A.~Yeung, ``Natural language processing
  as an emerging tool to detect late-life depression,'' \emph{Frontiers in
  Psychiatry}, vol.~12, 2021.

\bibitem{jackson2017naturala}
R.~G. Jackson, R.~Patel, N.~Jayatilleke, A.~Kolliakou, M.~Ball, G.~Gorrell,
  A.~Roberts, R.~J. Dobson, and R.~Stewart, ``Natural language processing to
  extract symptoms of severe mental illness from clinical text: The clinical
  record interactive search comprehensive data extraction ({{CRIS-CODE}})
  project,'' \emph{BMJ Open}, vol.~7, no.~1, p. e012012, Jan. 2017.

\bibitem{cohen2022integration}
J.~Cohen, J.~{Wright-Berryman}, L.~Rohlfs, D.~Trocinski, L.~Daniel, and T.~W.
  Klatt, ``Integration and validation of a natural language processing machine
  learning suicide risk prediction model based on open-ended interview language
  in the emergency department,'' \emph{Frontiers in Digital Health}, vol.~4,
  2022.

\bibitem{harvey2022natural}
D.~Harvey, F.~Lobban, P.~Rayson, A.~Warner, and S.~Jones, ``Natural language
  processing methods and bipolar disorder: Scoping review,'' \emph{JMIR Mental
  Health}, vol.~9, no.~4, p. e35928, Apr. 2022.

\bibitem{zhang2022natural}
T.~Zhang, A.~M. Schoene, S.~Ji, and S.~Ananiadou, ``Natural language processing
  applied to mental illness detection: A narrative review,'' \emph{npj Digital
  Medicine}, vol.~5, no.~1, pp. 1--13, Apr. 2022.

\bibitem{bedi2015automated}
G.~Bedi, F.~Carrillo, G.~A. Cecchi, D.~F. Slezak, M.~Sigman, N.~B. Mota,
  S.~Ribeiro, D.~C. Javitt, M.~Copelli, and C.~M. Corcoran, ``Automated
  analysis of free speech predicts psychosis onset in high-risk youths,''
  \emph{npj Schizophrenia}, vol.~1, no.~1, pp. 1--7, Aug. 2015.

\bibitem{calvo2017natural}
R.~A. Calvo, D.~N. Milne, M.~S. Hussain, and H.~Christensen, ``Natural language
  processing in mental health applications using non-clinical
  texts\textdagger,'' \emph{Natural Language Engineering}, vol.~23, no.~5, pp.
  649--685, Sep. 2017.

\bibitem{althoff2016largescale}
T.~Althoff, K.~Clark, and J.~Leskovec, ``Large-scale analysis of counseling
  conversations: An application of natural language processing to mental
  health,'' \emph{Transactions of the Association for Computational
  Linguistics}, vol.~4, pp. 463--476, 2016.

\bibitem{chattaraman2019should}
V.~Chattaraman, W.-S. Kwon, J.~E. Gilbert, and K.~Ross, ``Should {{AI-Based}},
  conversational digital assistants employ social- or task-oriented interaction
  style? {{A}} task-competency and reciprocity perspective for older adults,''
  \emph{Computers in Human Behavior}, vol.~90, pp. 315--330, Jan. 2019.

\bibitem{amin-nejad2020exploring}
A.~{Amin-Nejad}, J.~Ive, and S.~Velupillai,
  ``\BIBforeignlanguage{English}{Exploring transformer text generation for
  medical dataset augmentation},'' in
  \emph{\BIBforeignlanguage{English}{Proceedings of the 12th {{Language
  Resources}} and {{Evaluation Conference}}}}.\hskip 1em plus 0.5em minus
  0.4em\relax {Marseille, France}: {European Language Resources Association},
  May 2020, pp. 4699--4708.

\bibitem{ive2020generation}
J.~Ive, N.~Viani, J.~Kam, L.~Yin, S.~Verma, S.~Puntis, R.~N. Cardinal,
  A.~Roberts, R.~Stewart, and S.~Velupillai,
  ``\BIBforeignlanguage{en}{Generation and evaluation of artificial mental
  health records for natural language processing},''
  \emph{\BIBforeignlanguage{en}{npj Digital Medicine}}, vol.~3, no.~1, pp.
  1--9, May 2020.

\bibitem{soto2019neural}
X.~Soto, O.~{Perez-de-Vi{\~n}aspre}, G.~Labaka, and M.~Oronoz,
  ``\BIBforeignlanguage{en}{Neural machine translation of clinical texts
  between long distance languages},'' \emph{\BIBforeignlanguage{en}{Journal of
  the American Medical Informatics Association}}, vol.~26, no.~12, pp.
  1478--1487, Dec. 2019.

\bibitem{wolk2015neuralbased}
K.~Wo{\l}k and K.~Marasek, ``\BIBforeignlanguage{en}{Neural-based machine
  translation for medical text domain. based on european medicines agency
  leaflet texts},'' \emph{\BIBforeignlanguage{en}{Procedia Computer Science}},
  vol.~64, pp. 2--9, Jan. 2015.

\bibitem{wolk2016translation}
K.~Wolk and K.~P. Marasek, ``\BIBforeignlanguage{en}{Translation of medical
  texts using neural networks},'' \emph{\BIBforeignlanguage{en}{International
  Journal of Reliable and Quality E-Healthcare (IJRQEH)}}, vol.~5, no.~4, pp.
  51--66, 2016.

\end{thebibliography}

\begin{IEEEbiography}
    [{\includegraphics[width=1in,height=1.25in,clip,keepaspectratio]{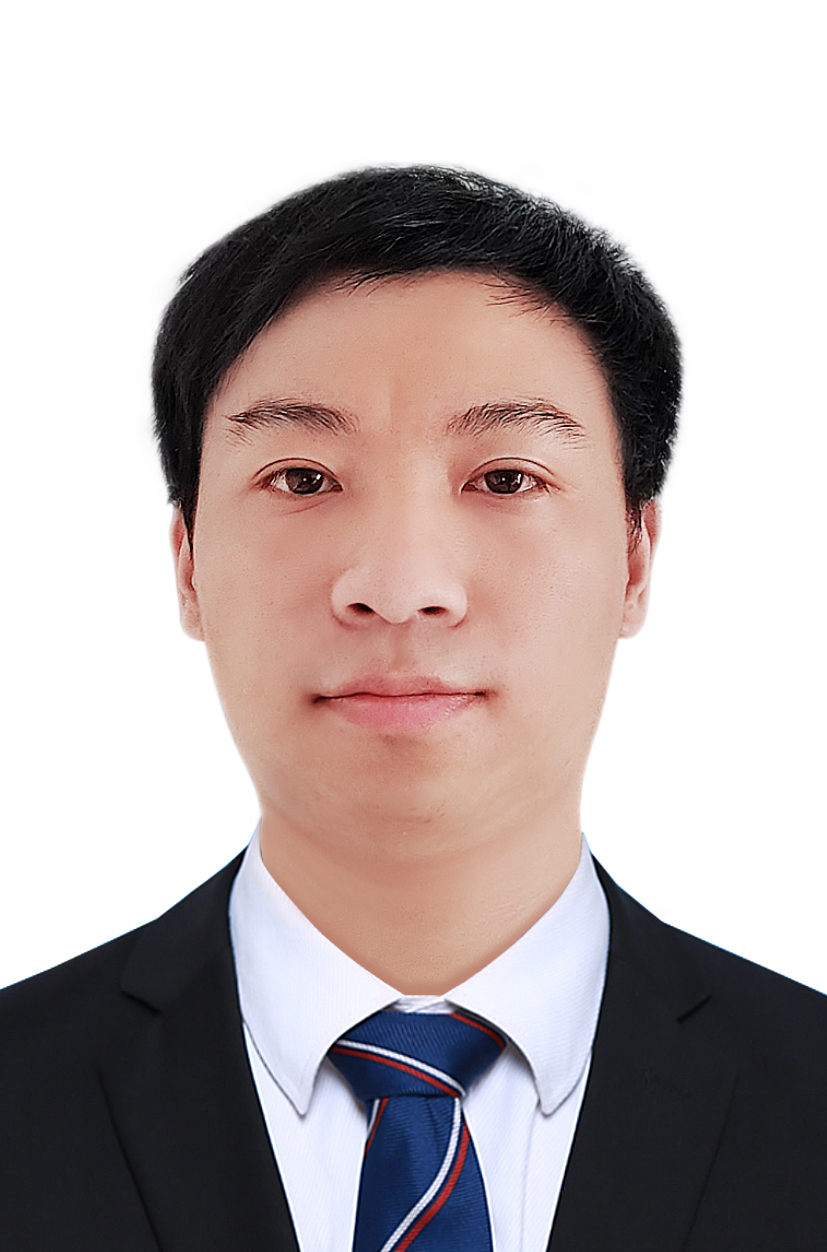}}]{Binggui Zhou} received the B.Eng. degree from Jinan University, Zhuhai, China, in 2018, and the M.Sc. degree from the University of Macau, Macao, China, in 2021, respectively. He is currently working toward the Ph.D. degree in Electrical and Computer Engineering with the University of Macau, Macao, China. He also serves as a Research Assistant with the School of Intelligent Systems Science and Engineering, Jinan University, Zhuhai, China. His research interests include Natural Language Processing, Artificial Intelligence, and AI assisted Wireless Communications.
\end{IEEEbiography}

\begin{IEEEbiography}
    [{\includegraphics[width=1in,height=1.25in,clip,keepaspectratio]{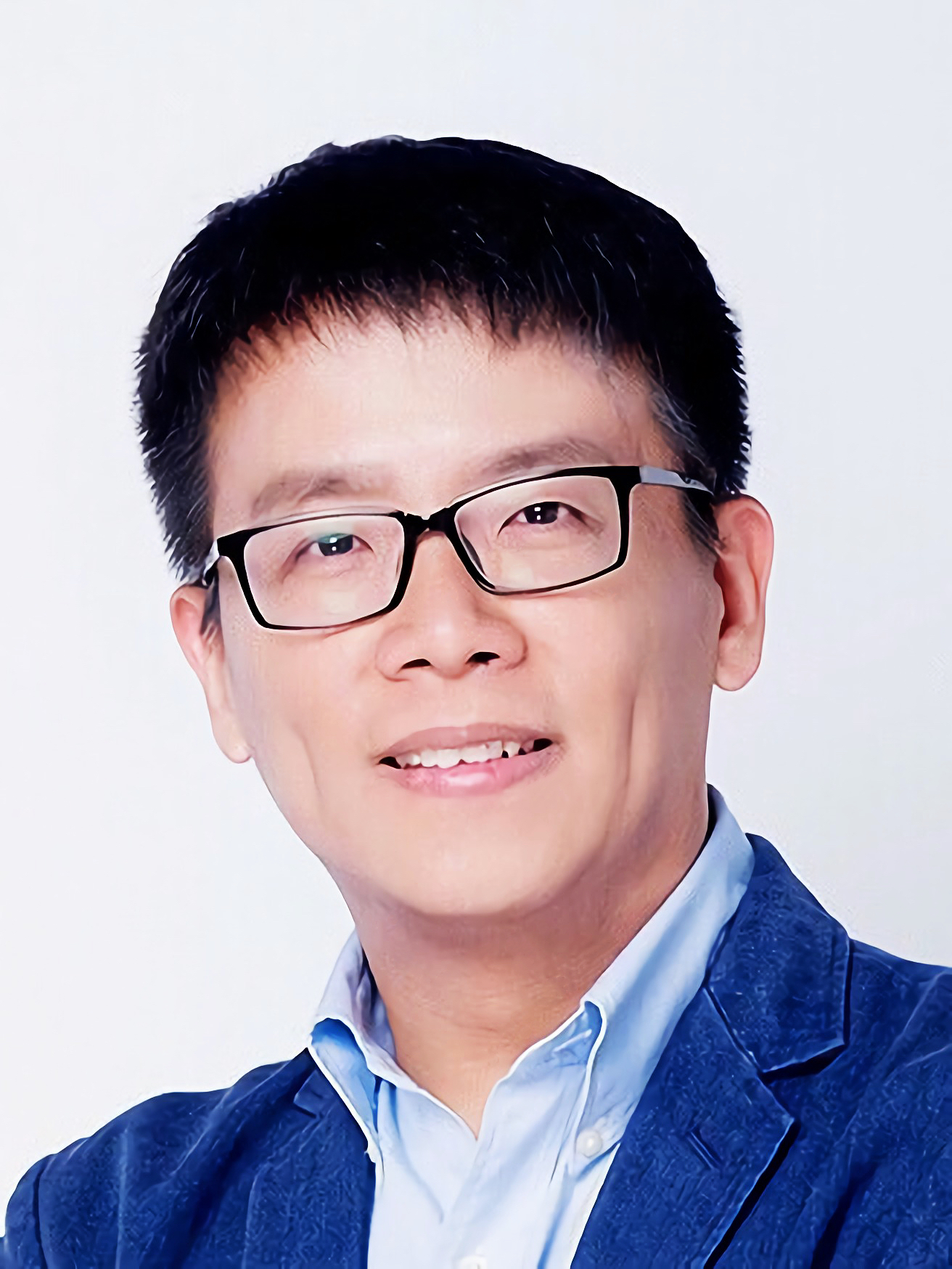}}]{Guanghua Yang} received his Ph.D. degree in electrical and electronic engineering from the University of Hong Kong, Hong Kong, in 2006. From 2006 to 2013, he served as post-doctoral fellow, research associate at the University of Hong Kong. Since April 2017, he has been with Jinan University, where he is currently a Full Professor in the School of Intelligent Systems Science and Engineering. His research interests are in the general areas of AI and its applications.
\end{IEEEbiography}

\begin{IEEEbiography}[{\includegraphics[width=1in,height=1.25in,clip,keepaspectratio]{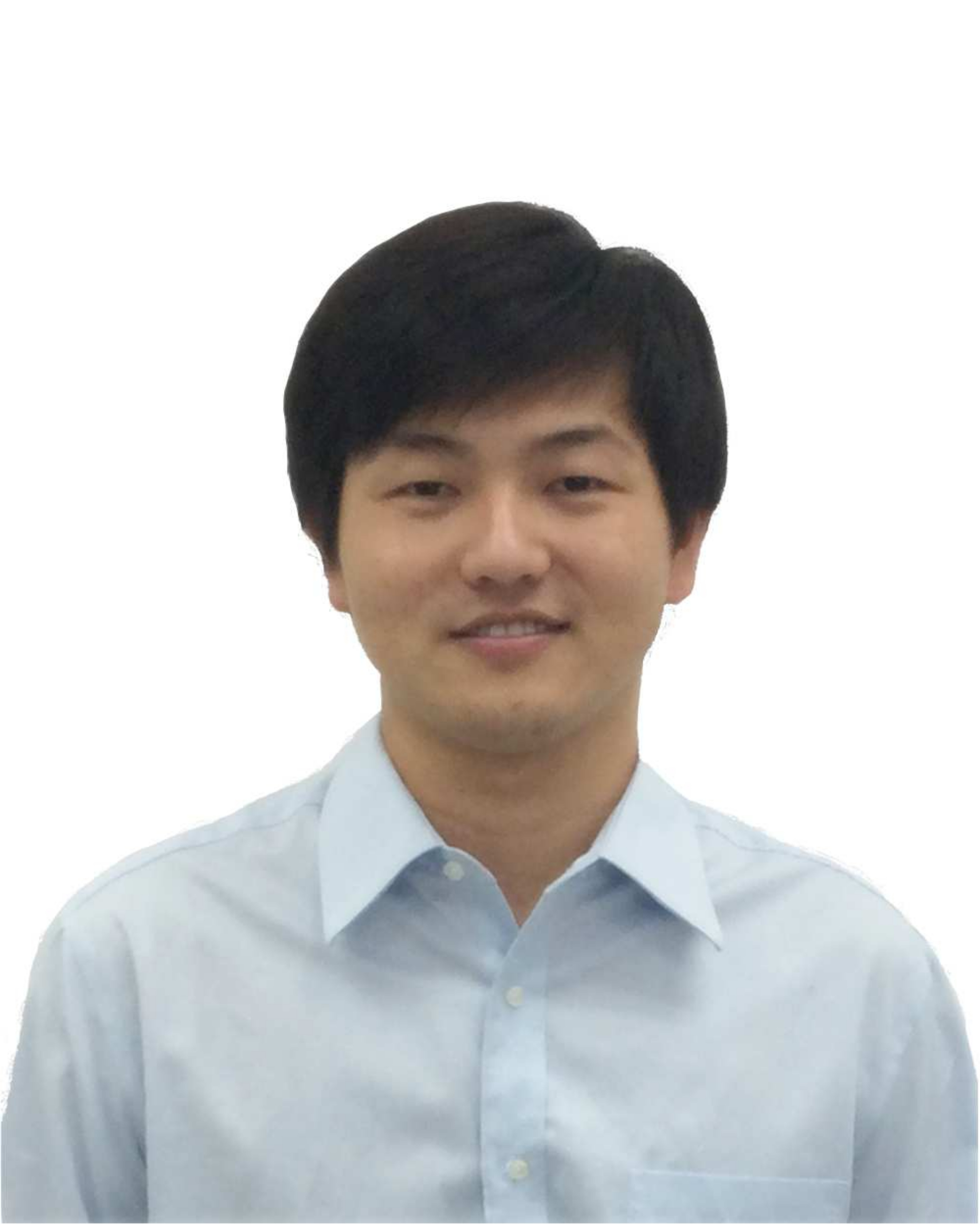}}]{Zheng Shi} received his B.S. degree in communication engineering from Anhui Normal University, China, in 2010 and his M.S. degree in communication and information system from Nanjing University of Posts and Telecommunications (NUPT), China, in 2013. He obtained his Ph.D. degree in Electrical and Computer Engineering from University of Macau, Macao, in 2017. He is currently an Associate Professor with the School of Intelligent Systems Science and Engineering, Jinan University, Zhuhai, China. His current research interests include hybrid automatic repeat request, non-orthogonal multiple access, machine learning and Internet of Things.
  \end{IEEEbiography}
  
\begin{IEEEbiography}
    [{\includegraphics[width=1in,height=1.25in,clip,keepaspectratio]{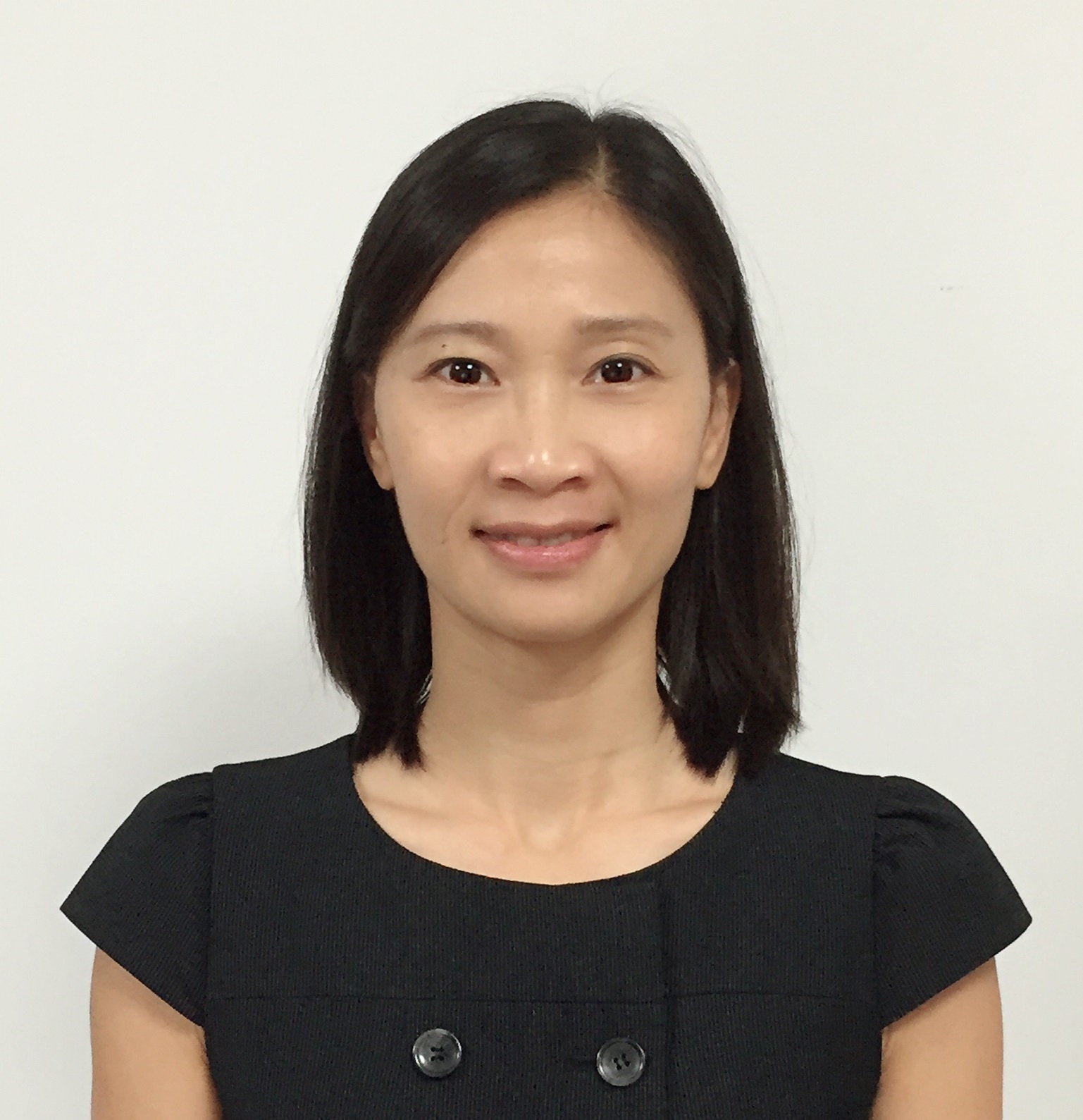}}]{Shaodan Ma} received the double Bachelor's degrees in science and economics and the M.Eng. degree in electronic engineering from Nankai University, Tianjin, China, in 1999 and 2002, respectively, and the Ph.D. degree in electrical and electronic engineering from The University of Hong Kong, Hong Kong, in 2006. From 2006 to 2011, she was a post-doctoral fellow at The University of Hong Kong. Since August 2011, she has been with the University of Macau, where she is currently a Professor. Her research interests include array signal processing, machine learning, wireless sensing and mmwave communications.
\end{IEEEbiography}

\end{document}